\documentclass[acmsmall,screen]{acmart}

\AtBeginDocument{%
  }

\setcopyright{none}
\acmDOI{}
\acmYear{2026}
\copyrightyear{2026}

\usepackage{amsmath}

\usepackage{amssymb}
\usepackage{amsthm}
\usepackage{algorithm}
\usepackage{algorithmic}
\usepackage{booktabs}
\usepackage{multirow}
\usepackage{dsfont}
\usepackage{url}
\usepackage[normalem]{ulem}
\useunder{\uline}{\ul}{}
\usepackage{subcaption}

\graphicspath{{./figures/}{./}}

\allowdisplaybreaks

\begin{document}

\title{Bayesian Tensor Autoencoder with Physics-informed Predictive Prior for Multi-dimensional Time Series Anomaly Detection}

\author{Jianan Liu}
\affiliation{%
  \institution{Zhejiang University}
  \city{Hangzhou}
  \country{China}}

\author{Chunguang Li}
\correspondingauthor
\affiliation{%
  \institution{Zhejiang University}
  \city{Hangzhou}
  \country{China}}
\email{cgli@zju.edu.cn}

\begin{abstract}
Multi-dimensional time series, inherently tensorial, are common in practice. Anomaly detection on such time series has a wide range of applications and the intrinsic correlations within these time series are beneficial for anomaly detection. Despite great progress in time series anomaly detection, most existing methods are conﬁned to uni-/multi-variate time series. When handling multi-dimensional time series using these methods, reshaping operations are required, which inevitably break the intrinsic correlations and thus lead to performance degradation. In uni-/multi-variate time series anomaly detection, AutoEncoders (AEs) are widely adopted and generally categorized into reconstruction-based and prediction-based AEs. The reconstruction-based AE utilizes the current observation for reconstruction, while the prediction-based AE utilizes the historical information to predict the current observation. Thus, the two AEs utilize different information. To bridge the gap between reconstruction-based and prediction-based AEs, so as to fully leverage the available information and thus further enhance performance, we propose a predictive prior and incorporate it into the reconstruction-based AE. It may not be very difficult to conceive this idea, but designing the predictive prior so that it can work for tensor anomaly detection is non-trivial. Specifically, to avoid breaking the intrinsic correlations within the multi-dimensional time series, we use the tensor AE as the backbone. To incorporate the predictive prior into the reconstruction-based AE, we propose a Bayesian fusion approach and our analysis reveals that this approach can enhance the modeling capability of the model for normal data. To mitigate the over-generalization problem of AE, we incorporate physical laws, i.e. tensor low-rank decomposition rules, into the neural networks in the predictive prior, leading to the Physics-informed Predictive Prior Tensor AE (PPPTAE) framework. Furthermore, training and testing strategies are tailored for PPPTAE so as to introduce sufficient randomness during training and bypass complex density estimation during testing, respectively. Experimental results on real-world datasets demonstrate the effectiveness of the proposed method.
\end{abstract}



\maketitle

\section{Introduction}

Multi-dimensional time series are common in practice. Take freeway traffic data as an example:  A freeway has multiple lanes, and sensors capture various road metrics along the freeway, with each sensor collecting simultaneous measurements for all lanes at the corresponding location over time. Such data are inherently tensorial and can be naturally represented as a fourth-order tensor sized $N_1 \times N_2 \times C \times T$, where $N_1$ is the number of sensors placed along the freeway, $N_2$ is the number of lanes, $C$ denotes the number of road metrics captured by sensors such as speed and occupancy, and $T$ is the total number of timestamps. Another example is the weather data, in which $N_1$ and $N_2$ denote the number of latitude and longitude points, respectively, and $C$ is the number of weather-related physical variables, such as temperature and humidity. In these examples, $N_1$ and $N_2$ contain location-related information and using a vector of length $N_1 \cdot N_2$ to represent the information would destroy the original location-related structure. Therefore, these data should be represented by tensors sized $N_1 \times N_2 \times C \times T$ so as to preserve the original structure.

Anomaly detection on multi-dimensional time series has a wide range of applications. For instance, detecting anomalies in freeway traffic data enables the prompt identification of abnormal events such as car accidents. Similarly, anomaly detection on weather data helps to timely detect extreme weather events such as tornadoes. By detecting anomalies, we can effectively minimize losses to life and property by promptly taking appropriate actions.

Different from supervised tasks such as time series classification \cite{classification1,classification2,classification3}, time series anomaly detection presents many unique challenges. On the one hand, as anomalies often correspond to rare events, normal samples often form a large portion of the data collected in practice, which leads to a highly imbalanced number of normal and abnormal samples in the real-world datasets. On the other hand, the types of anomalies are usually varied, making it extremely difficult to fully characterize abnormal events. In view of these problems, time series anomaly detection methods are often unsupervised and usually use only the normal samples to train the model so that normal patterns can be learned \cite{UAD1,UAD2,UAD3}. In this way, events that can not be described by the model are identified as anomalies during the testing stage.

Although time series anomaly detection is challenging, the intrinsic correlations within multi-dimensional time series can be beneficial for anomaly detection. On the one hand, by leveraging these intrinsic correlations, certain anomalies can be more effectively detected. For example, for car accidents on freeways, by integrating observation from multiple sensors located around the accident location, such abnormal events can be more effectively detected. On the other hand, leveraging the intrinsic correlations within multi-dimensional time series also helps in modeling normal data. For example, weather data present the nature that observation values at adjacent latitude and longitude points are similar though not identical. By taking the similarity into account, we can better model the normal data, thereby better distinguishing between normal and abnormal samples during the testing phase. Therefore, we should leverage the intrinsic correlations within multi-dimensional time series and not break these correlations.

Despite great progress in time series anomaly detection, most existing methods are conﬁned to uni-/multi-variate time series \cite{confined1,confined2,confined3}. When handling multi-dimensional time series using these methods, reshaping operations are required, which inevitably break the intrinsic correlations and thus lead to performance degradation.

In uni-/multi-variate time series anomaly detection, AutoEncoders (AEs) are widely adopted and generally categorized into reconstruction-based and prediction-based AEs. The reconstruction-based AE utilizes the current observation for reconstruction, while the prediction-based AE utilizes the historical information to predict the current observation. Thus, the two AEs utilize different information.

In order to bridge the gap between reconstruction-based and prediction-based AEs, so as to fully leverage the available information, thereby further enhancing performance, a natural idea is to concatenate the historical information and the current observation as the input of the model, and let the output be the current observation. Although this approach of concatenation is simple and intuitive, it fails to explicitly account for the interaction between historical information and current observations, and thus can not guarantee performance gains. To more effectively bridge the gap between the reconstruction-based AE and prediction-based AE, one possible way is to design a predictive prior using the historical information and incorporate this prior into the reconstruction-based AE, thereby effectively integrating prediction and reconstruction capabilities within a single unified model.

To the best of our knowledge, no existing work has adopted this idea. We note that the works known as the meta-analytic predictive prior \cite{pp1,pp2,pp3,pp4} have a similar name. These works assume that the parameters estimated from external studies follow the same distribution as those obtained in the current trial. Therefore, the distribution of parameters estimated from external data can be used as the prior for the parameters estimated in the current trial. However, this prior neglects temporal dependencies and uses a single global prior to model all external data. Therefore, the idea of the meta-analytic predictive prior is conceptually different from our idea and is unsuitable for time series tasks.

In this work, we propose a predictive prior and incorporate it into the reconstruction-based AE. It may not be very difficult to conceive this idea, but designing the predictive prior so that it can work for tensor anomaly detection is non-trivial. Specifically, to avoid breaking the intrinsic correlations within the multi-dimensional time series, we use the Tensor AE (TAE) as the backbone. To incorporate the predictive prior into the reconstruction-based TAE, we propose a Bayesian fusion approach and our analysis reveals that this approach can enhance the modeling capability of the model for normal data. To mitigate the over-generalization problem of AE, we incorporate physical laws, i.e. tensor low-rank decomposition rules, into the neural networks in the predictive prior, leading to the Physics-informed Predictive Prior Tensor AE (PPPTAE) framework. Furthermore, training and testing strategies are tailored for PPPTAE so as to introduce sufficient randomness during training and bypass complex density estimation during testing, respectively.

In summary, the contribution of this work is three-fold:
\begin{enumerate}
  \item We propose a predictive prior and incorporate it into the reconstruction-based TAE. In this way, we bridge the gap between prediction-based and reconstruction-based AEs, so as to fully leverage the available information, thereby further enhancing performance.
  \item To mitigate the over-generalization problem of AE, we incorporate physical laws, i.e. tensor low-rank decomposition rules, into the neural networks in the predictive prior, leading to the Physics-informed Predictive Prior Tensor AE (PPPTAE) framework.
  \item To incorporate the predictive prior into the reconstruction-based TAE, we propose a Bayesian fusion approach, and our analysis reveals that this approach can enhance the modeling capability of the model for normal data.
\end{enumerate}
\emph{Notations:} In this paper, we use lowercase boldface (e.g. $\mathbf{a}$), uppercase boldface (e.g. $\mathbf{A}$), and calligraphic letters (e.g. $\mathcal{A}$) to denote vectors, matrices, and tensors, respectively. We use $\mathbf{I}_{m}$ to denote the identity diagonal matrix of size $m \times m$. We use $[d] = \{ 1, 2, \cdots , d - 1, d \}$ to denote the set of integers from 1 to $d$, and use $a:b = \{a,a+1,...,b-1,b\}$ to denote the set of integers from $a$ to $b$. For a tensor $\mathcal{A} \in \mathbf{R}^{I_1 \times I_2 \times I_3}$, we use $\mathcal{A}_{i_1,i_2,i_3}$ to denote the $(i_1, i_2, i_3)$-th element of $\mathcal{A}$. The inner product of two tensors $\mathcal{A,B}$ is defined by $\langle \mathcal{A},\mathcal{B} \rangle = \sum_{i_1,i_2,i_3} \mathcal{A}_{i_1,i_2,i_3} \mathcal{B}_{i_1,i_2,i_3}$ and the squared Frobenius norm of a tensor $\mathcal{A}$ is defined by $\left\|\mathcal{A} \right\|_{F}^{2} = \langle \mathcal{A},\mathcal{A} \rangle$.

\section{Physics-informed Predictive Prior Tensor AutoEncoder}

In this section, we introduce the proposed Physics-informed Predictive Prior Tensor AutoEncoder (PPPTAE) framework.

Given multi-dimensional time series $\mathcal{X} \in \mathbf{R}^{N_1 \times N_2 \times C \times T}$, where $N_1$ and $N_2$ correspond to certain physical meanings (e.g., corresponding to the number of longitude and latitude points in weather data), $\mathcal{C}$ denotes the number of features, $T$ denotes the total number of timestamps. For the reconstruction-based Tensor AE (TAE), the input is the observation $ \mathcal{X}_{t} \in \mathbf{R}^{N_1 \times N_2 \times C}$ at timestamp $t$, and the corresponding feature $\mathcal{F}_{t} \in \mathbf{R}^{I_1 \times I_2 \times I_3}$ is extracted via the encoder. Subsequently, the decoder reconstructs the input according to the feature $\mathcal{F}_{t}$ to obtain $\hat{\mathcal{X}}_{t} \in \mathbf{R}^{N_1 \times N_2 \times C}$. During the testing stage, the deviation between the observation $ \mathcal{X}_{t}$ and its reconstruction $\hat{\mathcal{X}}_{t}$ is usually calculated to distinguish abnormal from normal samples. This is because during the training stage, only normal samples are used to train TAE, so the reconstruction error for normal samples is typically small, while the reconstruction error for abnormal samples is usually large. For the prediction-based TAE, the input is the observations from the previous $M$ timestamps $\mathcal{X}_{(t-M):(t-1)} \in \mathbf{R}^{N_1 \times N_2 \times C \times M}$. After feature extraction via the encoder, the decoder predicts the observation at timestamp $t$ and yields $\hat{\mathcal{X}}_{t}$. Consistent with the reconstruction-based AE, during the testing stage, the deviation between $ \mathcal{X}_{t}$ and $\hat{\mathcal{X}}_{t}$ is calculated to distinguish abnormal from normal samples.

To summarize, the reconstruction-based TAE utilizes the current observation for reconstruction, while the prediction-based TAE utilizes the historical information to predict the current observation. The two AEs utilize different information.

\subsection{Predictive Prior}

To bridge the gap between reconstruction-based and prediction-based TAEs, so as to fully leverage the available information, thereby further enhancing performance, we propose a predictive prior that can capture the temporal dependencies among features and incorporate it into the reconstruction-based TAE.

Specifically, to simplify the analysis and without loss of generality, we formulate the predictive prior as a Gaussian distribution. Furthermore, in order to leverage historical information to predict the mean and variance of this distribution, we construct functions $f_1(\cdot), f_2(\cdot)$ whose inputs are the $M$ features of the historical observations $\mathcal{F}_{t-1}, \mathcal{F}_{t-2},\cdots, \mathcal{F}_{t-M}$, and outputs are the mean and variance of the predictive prior, respectively. We formulate the proposed predictive prior as
\begin{align}
  p(\mathcal{F}_{t}\mid \mathcal{F}_{(t-M):(t-1)}) = \mathcal{N}(\mathcal{F}_{t};\left\langle\mathcal{F}_{t}\right\rangle,\sigma^{2}_{\mathcal{F}_{t}}) \label{pp}\\
	 \langle  \mathcal{F}_{t} \rangle = f_{1} (\mathcal{F}_{t-1}, \mathcal{F}_{t-2},\cdots, \mathcal{F}_{t-M}) \,\, \label{pp mean}\\
  \sigma^{2}_{\mathcal{F}_{t} } = f_{2} (\mathcal{F}_{t-1}, \mathcal{F}_{t-2},\cdots, \mathcal{F}_{t-M}), \label{pp var}
\end{align}
where $p(\cdot)$ is the probability density function, $\mathcal{N}(\cdot)$ is the Gaussian distribution, $\mathcal{F}_{t}$ denotes the feature tensor at timestamp $t$, $\left\langle\mathcal{F}_{t}\right\rangle$ denotes the mean of the feature tensor $\mathcal{F}_{t}$, the variance $\sigma^{2}_{\mathcal{F}_{t}}$ is element-wise, and  $\mathcal{F}_{t-1}, \mathcal{F}_{t-2},\cdots, \mathcal{F}_{t-M}$ are the $M$ features of the historical observations.

In the following, for clarity, we first introduce how to incorporate this predictive prior into the reconstruction-based TAE. Then we present the specific forms of the mappings $f_1(\cdot), f_2(\cdot)$.

\subsection{Bayesian Fusion Approach} \label{Bayes fusion}

We aim to utilize both the information from the $M$ features of the historical observations $\mathcal{F}_{(t-M):(t-1)}$, and the current observation $\mathcal{X}_t$, so as to obtain the current feature $\mathcal{F}_{t}$. In other words, our goal is to obtain the probability distribution $p(\mathcal{F}_t | \mathcal{X}_t, \mathcal{F}_{(t-M):(t-1)})$. To this end, we apply the Bayes' theorem to expand this probability distribution, leading to
\begin{align}
  p(\mathcal{F}_t |& \mathcal{X}_t, \mathcal{F}_{(t-M):(t-1)}) = \frac{p(\mathcal{X}_t | \mathcal{F}_t, \mathcal{F}_{(t-M):(t-1)}) \cdot p(\mathcal{F}_t | \mathcal{F}_{(t-M):(t-1)})}{p(\mathcal{X}_t | \mathcal{F}_{(t-M):(t-1)})}. \label{Bayes expand}
\end{align}
Since $p(\mathcal{X}_t | \mathcal{F}_t, \mathcal{F}_{(t-M):(t-1)})$ in the numerator of (\ref{Bayes expand}) models the conditional distribution of the observation given the features, it serves as a decoder. For the decoder of the reconstruction-based TAE, reconstructing the observation $\mathcal{X}_{t}$ only (directly) relies on its feature $\mathcal{F}_{t}$. Therefore, we have
\begin{equation}
  p(\mathcal{X}_t | \mathcal{F}_t, \mathcal{F}_{(t-M):(t-1)}) = p(\mathcal{X}_t | \mathcal{F}_t).
\end{equation}
We can thus further rewrite (\ref{Bayes expand}) as
\begin{align}
p(\mathcal{F}_t | \mathcal{X}_t, \mathcal{F}_{(t-M):(t-1)})  &= \frac{p(\mathcal{X}_t | \mathcal{F}_t) \cdot p(\mathcal{F}_t | \mathcal{F}_{(t-M):(t-1)})}{p(\mathcal{X}_t | \mathcal{F}_{(t-M):(t-1)})} = \frac{\frac{p( \mathcal{F}_t| \mathcal{X}_t)  p(\mathcal{X}_t) }{p(\mathcal{F}_t)}   \cdot p(\mathcal{F}_t | \mathcal{F}_{(t-M):(t-1)})}{p(\mathcal{X}_t | \mathcal{F}_{(t-M):(t-1)})} \notag \\
&= \frac{p( \mathcal{F}_t| \mathcal{X}_t)   \cdot p(\mathcal{F}_t | \mathcal{F}_{(t-M):(t-1)})}{\frac{p(\mathcal{F}_t)p(\mathcal{X}_t | \mathcal{F}_{(t-M):(t-1)})}{p(\mathcal{X}_t)}}.
\end{align}
Since $\mathcal{F}_{(t-M):(t-1)}$ and $\mathcal{X}_t$ are fixed when evaluating the distribution with respect to $\mathcal{F}_t$, both $p(\mathcal{X}_t | \mathcal{F}_{(t-M):(t-1)})$ and $p(\mathcal{X}_t)$ can be regarded as constants. We therefore have
\begin{equation}
  p(\mathcal{F}_t | \mathcal{X}_t, \mathcal{F}_{(t-M):(t-1)})= \alpha \cdot\frac{p( \mathcal{F}_t| \mathcal{X}_t)   \cdot p(\mathcal{F}_t | \mathcal{F}_{(t-M):(t-1)})}{p(\mathcal{F}_t)}, \label{posterior}
\end{equation}
where $\alpha = p(\mathcal{X}_t)/ p(\mathcal{X}_t | \mathcal{F}_{(t-M):(t-1)})$ is a constant. $p(\mathcal{F}_t| \mathcal{X}_t)$ is estimated by the encoder, which takes the observation $\mathcal{X}_{t}$ as input, and $p(\mathcal{F}_t | \mathcal{F}_{(t-M):(t-1)})$ is the predictive prior formulated in (\ref{pp}), which takes the features $\mathcal{F}_{(t-M):(t-1)}$ from the previous $M$ timestamps as input. Thus, (\ref{posterior}) is the Bayesian fusion formula.

For analytical tractability and without loss of generality, we model both $p(\mathcal{F}_t| \mathcal{X}_t)$ and $p(\mathcal{F}_t | \mathcal{F}_{(t-M):(t-1)})$ as Gaussian distributions, that is,
\begin{align}
p( \mathcal{F}_t| \mathcal{X}_t) &= \mathcal{N}( \mathcal{F}_t;\langle \mathcal{F}_{t}^{1} \rangle ,\sigma_{\mathcal{F}_{t}^{1}}^{2} ) \label{likelihood}\\
p(\mathcal{F}_t | \mathcal{F}_{(t-M):(t-1)}) &= \mathcal{N}( \mathcal{F}_t;\langle \mathcal{F}_{t}^{2} \rangle,\sigma_{\mathcal{F}_{t}^{2}}^{2} ). \label{prior}
\end{align}

For the numerator in (\ref{posterior}), the two Gaussian terms, $p( \mathcal{F}_t| \mathcal{X}_t)$ and $p(\mathcal{F}_t | \mathcal{F}_{(t-M):(t-1)})$, are multiplied together. Since the product of two Gaussian distributions is still Gaussian, the resulting distribution also follows a Gaussian form, whose mean and variance are given by:
\begin{align}
\mu_{\text{fused}} &= \frac{\lambda_1 \langle \mathcal{F}_{t}^{1} \rangle + \lambda_2 \langle \mathcal{F}_{t}^{2} \rangle}{\lambda_1 + \lambda_2} \label{mu_fuse}\\
\sigma_{\text{fused}}^2 &= \frac{1}{\lambda_1 + \lambda_2} \label{sigma_fuse},
\end{align}
where $\lambda_1 = 1/\sigma_{\mathcal{F}_{t}^{1}}^2$ and $ \lambda_2 = 1/\sigma_{\mathcal{F}_{t}^{2}}^2$ denote the precision of the two distributions in (\ref{likelihood}) and (\ref{prior}), respectively.

For the denominator $p(\mathcal{F}_{t})$ in (\ref{posterior}), by the rules of probability, this density satisfies
\begin{equation}
  p(\mathcal{F}_t) = \int p(\mathcal{F}_t | \mathcal{X}_{t}) p(\mathcal{X}_{t}) d\mathcal{X}_{t}, \label{int Ft}
\end{equation}
where $p(\mathcal{X}_{t})$ is the probability density of the training data.

To demonstrate the effect of the denominator $p(\mathcal{F}_{t})$ in (\ref{posterior}), we further define
\begin{equation}
  p(\mathcal{F}_t^{\text{pos}}) \triangleq \int  p(\mathcal{F}_t | \mathcal{X}_t, \mathcal{F}_{(t-M):(t-1)}) p(\mathcal{X}_{t})d\mathcal{X}_{t}  \label{p(Fpos)}.
\end{equation}
Note that $\mathcal{F}_{(t-M):(t-1)}$ is already determined at timestamp $t$, so no additional integration over $\mathcal{F}_{(t-M):(t-1)}$ is required. When substituting (\ref{posterior}) and (\ref{int Ft}) into (\ref{p(Fpos)}), we obtain
\begin{align}
p(\mathcal{F}_t^{\text{pos}}) &\propto   \int  \frac{p( \mathcal{F}_t| \mathcal{X}_t)   \cdot p(\mathcal{F}_t | \mathcal{F}_{(t-M):(t-1)})}{p(\mathcal{F}_t)}p(\mathcal{X}_{t})d\mathcal{X}_{t} \notag \\
&=   \frac{\int p( \mathcal{F}_t| \mathcal{X}_t)   \cdot p(\mathcal{F}_t | \mathcal{F}_{(t-M):(t-1)})p(\mathcal{X}_{t})d\mathcal{X}_{t}}{p(\mathcal{F}_t)} \notag \\
&=   \frac{\int p( \mathcal{F}_t| \mathcal{X}_t)   \cdot p(\mathcal{F}_t | \mathcal{F}_{(t-M):(t-1)})p(\mathcal{X}_{t})d\mathcal{X}_{t}}{\int p(\mathcal{F}_t | \mathcal{X}_{t}) p(\mathcal{X}_{t}) d\mathcal{X}_{t}} \notag \\
&\approx \frac{\sum_{i} p( \mathcal{F}_t| \mathcal{X}_t^{(i)})   \cdot p(\mathcal{F}_t | \mathcal{F}_{(t-M):(t-1)}^{(i)})}{\sum_{i} p(\mathcal{F}_t | \mathcal{X}_{t}^{(i)})}, \label{uniform}
\end{align}
where $\mathcal{X}_{t}^{(i)} \sim p(\mathcal{X}_{t})$ denotes a training sample drawn from the training data distribution.

\begin{figure}[tbp]  
  \centering  
  \includegraphics[width=8cm]{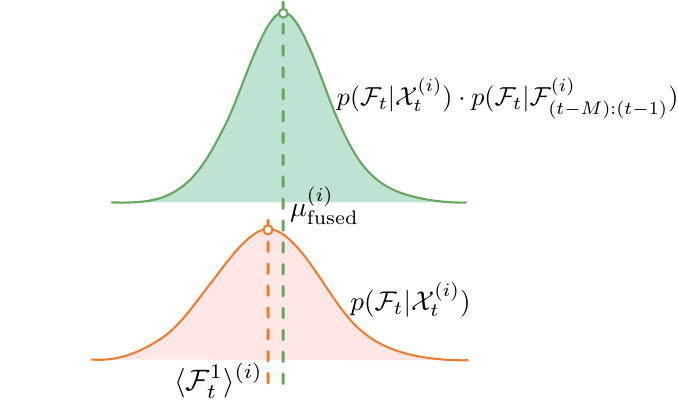} 
  \caption{A schematic illustration of the distributions $p(\mathcal{F}_t \mid \mathcal{X}_t^{(i)}) \cdot p(\mathcal{F}_t \mid \mathcal{F}_{(t-M):(t-1)}^{(i)})$ and $p(\mathcal{F}_t \mid \mathcal{X}_t^{(i)})$, where $\mathcal{X}_t^{(i)} \sim p(\mathcal{X}_t)$ is a training sample drawn from the data distribution. Since the mean $\mu_{\text{fused}}^{(i)}$ of $p(\mathcal{F}_t \mid \mathcal{X}_t^{(i)}) \cdot p(\mathcal{F}_t \mid \mathcal{F}_{(t-M):(t-1)}^{(i)})$ is close to the mean $\langle \mathcal{F}_{t}^{1}\rangle^{(i)}$ of $p(\mathcal{F}_t \mid \mathcal{X}_t^{(i)})$, the probability densities of the two distributions exhibit an approximately positive correlation. That is, when the probability density of $p(\mathcal{F}_t \mid \mathcal{X}_t^{(i)}) \cdot p(\mathcal{F}_t \mid \mathcal{F}_{(t-M):(t-1)}^{(i)})$ is large, $p(\mathcal{F}_t \mid \mathcal{X}_t^{(i)})$ is roughly also large, and vice versa.}  
  \Description{A schematic plot comparing two related probability density functions over feature values, illustrating that their densities are approximately positively correlated.}
  \label{uniform_fig}
\end{figure}

For the $i$-th distribution of the numerator in (\ref{uniform}), the product of the two distributions remains Gaussian, with mean $\mu_{\text{fused}}^{(i)}$ given in (\ref{mu_fuse}). Besides, since the model is trained only on normal samples, which are generally predictable, and the predictive prior and the encoder should perform well after training, $\mu_{\text{fused}}^{(i)}$ would be around $\langle \mathcal{F}_{t}^{1}\rangle^{(i)}$, which is the mean of the $i$-th component in the denominator of (\ref{uniform}). Consequently, as illustrated in Fig. \ref{uniform_fig}, the probability densities of the $i$-th numerator and denominator exhibit an approximately positive correlation. That is, when the probability density of the numerator is large, the denominator is roughly also large, and vice versa. Consequently, the ratio of their probability densities fluctuates within a narrower range, thereby becoming more concentrated. When summing over all distributions, the same relationship between the numerator and denominator remains, so $p(\mathcal{F}_t^{\text{pos}}) $ exhibits a tendency toward a uniform distribution. Under this effect, the resulting features $\mathcal{F}_t^{\text{pos}}$, which are fed into the decoder for reconstruction, become more evenly distributed, thereby allowing each region in the feature space to contain sufficient training samples. As a result, the reconstruction capability corresponding to each region is well established, thereby preventing reconstruction degradation in regions with insufficient features. Thus, during the testing stage, normal samples are reconstructed consistently well. In summary, the denominator $p(\mathcal{F}_{t})$ in (\ref{posterior}) functions as a reweighting factor and can provide a beneficial contribution to modeling normal data.



So far, we have incorporated the predictive prior into the reconstruction-based TAE using the Bayesian approach. Next, we present the specific forms of the mappings $f_1(\cdot), f_2(\cdot)$.

\subsection{Physics-informed Predictive Prior}

For $f_1(\cdot),f_2(\cdot)$, inspired by the universal approximation theorem \cite{universal} of neural networks, a natural idea is to directly construct two neural networks to achieve prediction, that is,
\begin{align}
  \langle  \mathcal{F}_{t} \rangle &= \text{NN}_{1} (\mathcal{F}_{t-1}, \mathcal{F}_{t-2},\cdots, \mathcal{F}_{t-M})\\
  \sigma^{2}_{\mathcal{F}_{t} } &= \text{NN}_{2} (\mathcal{F}_{t-1}, \mathcal{F}_{t-2},\cdots, \mathcal{F}_{t-M}),
\end{align}
where $\mathcal{F}_{t-1}, \mathcal{F}_{t-2},\cdots, \mathcal{F}_{t-M}$ are the $M$ features of the historical observations, $\left\langle\mathcal{F}_{t}\right\rangle$ and $\sigma^{2}_{\mathcal{F}_{t}}$ are the mean and variance of the predictive prior distribution of the feature $\mathcal{F}_{t}$ at timestamp $t$, respectively. However, this approach cannot mitigate the over-generalization problem of reconstruction-based AE, i.e., sometimes abnormal samples may also be reconstructed well \cite{overgen1,overgen2,overgen3}. To this end, we introduce a Bayesian tensor decomposition module in $f_1(\cdot), f_2(\cdot)$ so as to impose a low-rank constraint on the normal features.

In the following, we first introduce the details of the Bayesian tensor decomposition module, and then we carefully analyze its mechanism for mitigating the over-generalization problem of AE. Finally, we present the overall design of the physics-informed predictive prior.

\subsubsection{Bayesian Tensor Decomposition Module} \label{Bayesian TD}

For the features of $M$ historical observations, $\mathcal{F}_{t-1},\\ \mathcal{F}_{t-2}, \cdots, \mathcal{F}_{t-M}$, we first decompose them respectively using the Bayesian tensor decomposition module. Specifically, similar to previous tensor decomposition frameworks \cite{framework1,framework2,framework3}, we also assume that the feature $\mathcal{F}$ can be decomposed into a low-rank component $\mathcal{L}$, a sparse component $\mathcal{S}$ and a noise component $\mathcal{E}$, that is,
\begin{equation}
  \mathcal{F} = \mathcal{L} + \mathcal{S} + \mathcal{E} \label{decomp},
\end{equation}
where the low-rank component $\mathcal{L}$ captures normal features. Since the training process involves only normal samples, using $\mathcal{L}$ as the input to the decoder is sufficient for accurate reconstruction. Therefore, our objective is to estimate $\mathcal{L}$ from $\mathcal{F}$. To this end, we first specify the low-rank structure of $\mathcal{L}$.

Recently, many novel low-rank structures have been proposed, such as tensor train \cite{TT1,TT2}, tensor ring \cite{TR1,TR2}, and tensor Singular Value Decomposition (t-SVD) \cite{tsvd1,tsvd2}. In comparison to these low-rank structures, the Tensor Wheel (TW) model can comprehensively establish all possible mode interactions within a tensor, thereby more accurately capturing the complex interactions within the tensor \cite{TW}. Here, we choose to use the TW model to characterize the low-rank structure of the feature $\mathcal{F}$, that is,
\begin{equation}
  \mathcal{L} =\mathrm{TW} [\{\mathcal{G}^{(k)}\}_{k \in [3]} ; \mathcal{C}], \label{TW model}
\end{equation}
where $\mathcal{G}^{(k)} \in \mathbf{R}^{R_{k} \times R_{k+1} \times L_{k} \times I_{k}}, k \in [3]$ are the TW ring factors, and when $k=3$, $R_4$ stands for $R_1$; $\mathcal{C} \in \mathbf{R}^{L_1 \times L_2 \times  L_3}$ is the TW core factor, and $\mathbf{r} =(R_1,R_2,R_3,L_1,L_2,L_3) \in \mathbf{R}^{6} $ is defined as the TW rank. 

\begin{figure}[tbp]  
  \centering  
  \includegraphics[width=8cm]{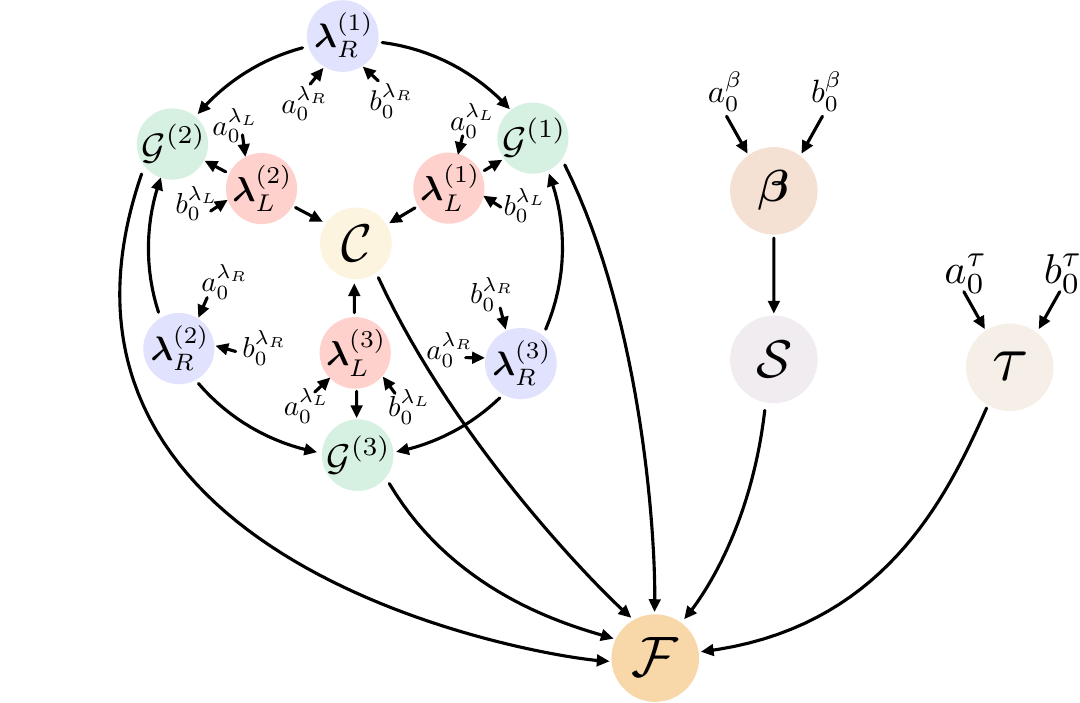} 
  \caption{The graphical model of the Bayesian tensor decomposition model.}  
  \Description{A probabilistic graphical model showing the dependencies among tensor factors, core tensor, sparse component, noise precision, and observed feature tensor in the Bayesian tensor decomposition model.}
  \label{graph}
\end{figure}

Since we use the TW model to characterize the low-rank structure of $\mathcal{L}$, estimating $\mathcal{L}$ reduces to estimating the TW ring factors $\{\mathcal{G}^{(k)}\}_{k \in [3]}$ and the TW core factor $\mathcal{C}$. In addition, to obtain an accurate estimation of $\mathcal{L}$, it is also necessary to model the sparse component $\mathcal{S}$ and the noise component $\mathcal{E}$. Accordingly, probabilistic formulations are introduced for $\mathcal{L}$, $\mathcal{S}$, and $\mathcal{E}$ to maintain compatibility with the proposed predictive prior and facilitate effective inference. For brevity, the detailed probabilistic formulations are provided in Appendix \ref{probabilistic Modeling}, while Fig. \ref{graph} illustrates the corresponding graphical model and the resulting joint distribution is given by
\begin{align}
p(\mathcal{F},\boldsymbol{\Theta}) = & p(\mathcal{F} \mid \{\mathcal{G}^{(k)}\}_{k \in [3]},\mathcal{C},\mathcal{S},\tau)  \prod_{k=1}^{3} p\left(\mathcal{G}^{(k)}\mid \boldsymbol{\lambda}^{(k)}_{R}, \boldsymbol{\lambda}^{(k+1)}_{R}, \boldsymbol{\lambda}^{(k)}_{L}\right) \notag \\
& \times p(\mathcal{C} \mid \{\boldsymbol{\lambda}_{L}^{(k)}\}_{k \in [3]} )\prod_{k=1}^{3} p(\boldsymbol{\lambda}_R^{(k)})  \prod_{k=1}^{3} p(\boldsymbol{\lambda}_L^{(k)})p(\mathcal{S} \mid \boldsymbol{\beta}) p(\boldsymbol{\beta})p(\tau),
\end{align}
where  $\{\boldsymbol{\lambda}^{(k)}_{R}\}_{k \in [3]}$ and $\{\boldsymbol{\lambda}^{(k)}_{L}\}_{k \in [3]}$ are the precision parameters associated with the TW ring factors and the TW core factor, respectively; $\boldsymbol{\beta}$ is the precision parameter associated with the sparse component; $\tau$ is the precision parameter associated with the noise component; $\boldsymbol{\Theta} = \{ \{\mathcal{G}^{(k)}\}_{k \in [3]},\mathcal{C},\\\{\boldsymbol{\lambda}^{(k)}_{R}\}_{k \in [3]},\{\boldsymbol{\lambda}^{(k)}_{L}\}_{k \in [3]}, \mathcal{S},\boldsymbol{\beta},\tau\}$ is the set of all variables.

To infer the posterior distribution of the variables given the feature $\mathcal{F}$, i.e., $p(\boldsymbol{\Theta} \mid \mathcal{F})$, according to the Bayesian rule, we theoretically have
\begin{equation}
  p(\boldsymbol{\Theta} \mid \mathcal{F}) = \frac{p(\mathcal{F},\boldsymbol{\Theta})}{\int p(\mathcal{F},\boldsymbol{\Theta})d \boldsymbol{\Theta}}.
\end{equation}
However, as some of the variables are conditionally dependent, the marginal distribution in the denominator is analytically intractable, impeding the computation of the posterior distribution. Therefore, we resort to the Variational Inference (VI) method \cite{VB1,VB2} to approximate the posterior distribution. Specifically, under the commonly used mean-field assumption, we assume the variational posterior to be
\begin{align}
  q(\boldsymbol{\Theta})=q(\tau) \prod_{k=1}^{3} q(\mathcal{G}^{(k)}) q(\mathcal{C}) \prod_{k=1}^{3} q(\boldsymbol{\lambda}_R^{(k)})  \prod_{k=1}^{3} q(\boldsymbol{\lambda}_L^{(k)}) q(\mathcal{S}) q(\boldsymbol{\beta}), \label{VB pos}
\end{align}
and our objective is to minimize the KL divergence between the variational posterior and the true posterior, i.e.
\begin{equation}
  \min \mathrm{KL}[q(\boldsymbol{\Theta}) || p(\boldsymbol{\Theta} \mid \mathcal{F})], \label{KL}
\end{equation}

According to the VI theory, as all variables are assigned conjugate priors, the optimal variational posterior distributions admit closed-form expressions, which are provided in Appendix \ref{optimal}.

Although the VI algorithm can effectively solve the optimization problem (\ref{KL}) so that the variational posterior can approximate the true posterior distribution well, the VI algorithm requires many steps of iterations, resulting in high computational complexity. To mitigate this issue, inspired by the strengths of neural networks in learning complex nonlinear mapping, we design neural networks to achieve Bayesian tensor decomposition, thereby reducing the computational complexity. Specifically, we design neural networks to learn the nonlinear mapping between the feature $\mathcal{F}$ and the variational posterior parameters of the TW ring factors $\{\mathcal{G}^{(k)}\}_{k \in [3]}$, the TW core factor $\mathcal{C}$, and the sparse component $\mathcal{S}$, i.e.,
\begin{align}
[\langle \mathcal{G}^{(k)}\rangle, \sigma_{\mathcal{G}^{(k)}}^{2} ]&=f_{\mathcal{G}^{(k)}}(\mathcal{F}), k \in[3] \label{output G}\\
[\langle \mathcal{C}\rangle, \sigma_{\mathcal{C}}^{2}]&=f_{\mathcal{C}}(\mathcal{F}) \label{output C}\\
[\langle \mathcal{S}\rangle, \sigma_{\mathcal{S}}^{2}]&=f_{\mathcal{S}}(\mathcal{F}), \label{output S}
\end{align}
where the nonlinear functions $f_{\mathcal{G}^{(k)}}(\cdot),f_{\mathcal{C}}(\cdot),f_{\mathcal{S}}(\cdot)$ are all parameterized by neural networks.

At first glance, the design of these networks seems straightforward. However, since their outputs are parameters of probabilistic distributions, formulating appropriate objective functions to effectively train them is non-trivial. In the following, we present the design of the objective function.

To design an objective function to train these neural networks, we first convert the KL divergence in (\ref{KL}) as follows:
\begin{align}
\mathrm{KL}[q(\boldsymbol{\Theta}) || p(\boldsymbol{\Theta} \mid \mathcal{F})] &= \mathbf{E}[ \ln q(\boldsymbol{\Theta})] - \mathbf{E}[\ln p(\boldsymbol{\Theta} \mid \mathcal{F})]  \notag\\
&=\mathrm{KL}[q(\boldsymbol{\Theta}) || p(\boldsymbol{\Theta})] - \mathbf{E}[\ln p(\mathcal{F} \mid \boldsymbol{\Theta})]  + \ln p(\mathcal{F}), \label{KL expand}
\end{align}
where all expectations are with respect to $q(\boldsymbol{\Theta})$, and the variable to be optimized is $\boldsymbol{\Theta}$.

Since $\ln p(\mathcal{F})$ is independent to the variable $\boldsymbol{\Theta}$, $\ln p(\mathcal{F})$ can be viewed as a constant when optimizing the variable $\boldsymbol{\Theta}$, that is,
\begin{align}
\min_{q(\boldsymbol{\Theta})} \mathrm{KL}[q(\boldsymbol{\Theta}) || p(\boldsymbol{\Theta} \mid \mathcal{F})] \Leftrightarrow  \min_{q(\boldsymbol{\Theta})} \mathrm{KL}[q(\boldsymbol{\Theta}) || p(\boldsymbol{\Theta})] - \mathbf{E}[\ln p(\mathcal{F} \mid \boldsymbol{\Theta})]. \label{eq.5}
\end{align}

\begin{table}[t]
\centering
\caption{The expressions and interpretations for items in the VB loss function $L_{\text{VB}}$ (\ref{loss VB})}
\label{tab:loss VB}
\resizebox{\textwidth}{!}{%
\begin{tabular}{@{}c|c@{}}
\toprule
Item &
  Expression and Interpretation \\ \toprule
\multirow{2}{*}{$L_{\mathcal{F}}$} &
  $L_{\mathcal{F}} =   \frac{1}{2} \langle\tau\rangle\left\|\mathcal{F} - \mathrm{TW}   [\{\mathcal{G}^{(k)}\}_{k \in [3]} ; \mathcal{C}] - \mathcal{S}   \right\|^{2}_{F}$ \\ \cmidrule(l){2-2} 
 &
  Ensure the consistency between feature $\mathcal{F}$ and the tensor decomposition results, $\mathrm{TW}   [\{\mathcal{G}^{(k)}\}_{k \in [3]} ; \mathcal{C}]$ and $\mathcal{S}$. \\ \bottomrule
\multirow{2}{*}{$L_{\langle \mathcal{G}^{(k)}\rangle}$} &
  $L_{\langle  \mathcal{G}^{(k)}\rangle} = \frac{1}{2} \sum_{r_k=1}^{R_k}   \sum_{r_{k+1}=1}^{R_{k+1}} \sum_{l_k =1}^{L_k} \sum_{i_k =1}^{I_k}   \langle\lambda_{r_k}^{(k)}\rangle \langle \lambda_{r_{k+1}}^{(k+1)} \rangle   \langle\lambda_{l_k}^{(k)} \rangle \left\langle\mathcal{G}^{(k)}(r_k,r_{k+1},l_k,i_k)\right\rangle^{2}$ \\ \cmidrule(l){2-2} 
 &
  Adapt to the corresponding elements of the TW rank. \\ \bottomrule
\multirow{2}{*}{$L_{\sigma_{g^{k}}^{2}}$} &
  $L_{\sigma_{g^{k}}^{2}} =   \frac{1}{2}\sum_{r_k=1}^{R_k} \sum_{r_{k+1}=1}^{R_{k+1}} \sum_{l_k =1}^{L_k}   \sum_{i_k =1}^{I_k} (\langle\lambda_{r_k}^{(k)}\rangle \langle   \lambda_{r_{k+1}}^{(k+1)} \rangle \langle\lambda_{l_k}^{(k)}   \rangle\sigma_{g^{k}_{r_k,r_{k+1},l_k,i_k}}^{2} - \ln   \sigma_{g^{k}_{r_k,r_{k+1},l_k,i_k}}^{2} ) $ \\ \cmidrule(l){2-2} 
 &
  Prevent the variance from shrinking to zero, so as to prevent $q(\mathcal{G}^{(k)})$ from degenerating into a one-point distribution. \\ \bottomrule
\multirow{2}{*}{$L_{\langle \mathcal{C} \rangle}$} &
  $L_{\langle \mathcal{C}   \rangle} = \frac{1}{2} \sum_{l_1 l_2 l_3}\langle\lambda_{l_1}^{(1)}   \rangle\langle\lambda_{l_2}^{(2)} \rangle \langle\lambda_{l_3}^{(3)}\rangle   \left\langle\mathcal{C}_{l_1 l_2 l_3}\right\rangle^{2}$ \\ \cmidrule(l){2-2} 
 &
  Adapt to the corresponding elements of the TW rank. \\ \bottomrule
\multirow{2}{*}{$L_{\sigma_{C}^{2}}$} &
  $L_{\sigma_{C}^{2}} =   \frac{1}{2}\sum_{l_1 l_2 l_3}(\langle\lambda_{l_1}^{(1)}   \rangle\langle\lambda_{l_2}^{(2)} \rangle   \langle\lambda_{l_3}^{(3)}\rangle\sigma_{C_{l_1 l_2 l_3}}^{2} - \ln   \sigma_{C_{l_1 l_2 l_3}}^{2} )$ \\ \cmidrule(l){2-2} 
 &
  Prevent the variance from shrinking to zero, so as to prevent $q(\mathcal{C})$ from degenerating into a one-point distribution. \\ \bottomrule
\multirow{2}{*}{$L_{\langle   \mathcal{S}\rangle}$} &
  $L_{\langle   \mathcal{S}\rangle} = \frac{1}{2} \sum_{i_1 i_2 i_3 } \langle \beta_{i_1 i_2   i_3} \rangle \left\langle \mathcal{S}_{i_1 i_2 i_3}\right\rangle^{2}$ \\ \cmidrule(l){2-2} 
 &
  Encourage $\langle \mathcal{S} \rangle$ to be sparse. \\ \bottomrule
\multirow{2}{*}{$L_{\sigma_{S}^{2}}$} &
  $L_{\sigma_{S}^{2}} =   \frac{1}{2} \sum_{i_1 i_2 i_3 }( \langle \beta_{i_1 i_2 i_3} \rangle   \sigma^{2}_{S_{i_1 i_2 i_3}} - \ln \sigma^{2}_{S_{i_1 i_2 i_3}} )$ \\ \cmidrule(l){2-2} 
 &
  Prevent the variance from shrinking to zero, so as to prevent $q(\mathcal{S})$ from degenerating into a one-point distribution. \\ \bottomrule
\end{tabular}%
}
\end{table}

Next, we derive the two terms on the right-hand side of (\ref{eq.5}) to obtain the objective function. For clarity, we begin with the second term. According to the conditional distribution of the feature $\mathcal{F}$ (see (\ref{F likelihood}) in Appendix \ref{probabilistic Modeling}), the second term on the right-hand side of (\ref{eq.5}) can be written as:
\begin{align}
\mathbf{E}&[\ln p(\mathcal{F} \mid \boldsymbol{\Theta})]=\sum_{i_1=1}^{I_1} \sum_{i_2=1}^{I_2}\sum_{i_3 =1}^{I_3} \left\langle \frac{1}{2} \ln \tau - \frac{\tau}{2} \Bigl(  \mathcal{F}_{i_1 i_2 i_3} -  \mathrm{TW} [\{\mathcal{G}^{(k)}\}_{k\in[3]} ; \mathcal{C}]_{i_1 i_2 i_3} - \mathcal{S}_{i_1 i_2 i_3}\Bigl)^{2} \right\rangle _{q(\boldsymbol{\Theta})}. \label{eq.6}
\end{align}
Due to the presence of quadratic terms, the calculation of the expectation becomes complicated. To this end, we adopt the reparameterization trick \cite{repara1,repara2} to circumvent the direct calculation of the complicated expectation terms. Specifically, let $\epsilon$ denote a random noise variable drawn from the standard normal distribution, i.e., $\epsilon \sim \mathcal{N}(0,1)$. Then, we have
\begin{align}
\mathcal{G}^{(k)} &= \langle \mathcal{G}^{(k)}\rangle + \sigma_{g^{k}} \odot \boldsymbol{\epsilon }, k \in [3] \label{repa1}\\
\mathcal{C} &= \langle  \mathcal{C}  \rangle + \sigma_{C} \odot \boldsymbol{\epsilon } \label{repa2}\\
\mathcal{S} &= \langle  \mathcal{S}  \rangle + \sigma_{S} \odot \boldsymbol{\epsilon }, \label{repa3}
\end{align}
where $\odot$ denotes the element-wise multiplication. Therefore, (\ref{eq.6}) can be further converted to
\begin{align}
  \mathbf{E}[\ln p(\mathcal{F} \mid \boldsymbol{\Theta})] &= \mathbf{E}_{q(\tau)}[\mathbf{E}_{q(\{\mathcal{G}^{(k)}\}_{k\in [3]}, \mathcal{C},\mathcal{S})}[\ln p(\mathcal{F} \mid \boldsymbol{\Theta})]] \notag\\
  &\approx \mathbf{E}_{q(\tau)}\left[\frac{1}{N} \sum_{i=1}^{N} \ln p(\mathcal{F} \mid \{\mathcal{G}_i^{(k)}\}_{k \in [3]}, \mathcal{C}_i,\mathcal{S}_i ,\tau)\right],
\end{align}
where $\left\{\{\mathcal{G}_i^{(k)}\}_{k \in [3]}, \mathcal{C}_i,\mathcal{S}_i\right\}_{i=1}^{N}$ are the reparameterized samples. In practice, $N$ is typically set to 1, so that
\begin{align}
  \mathbf{E}[\ln p(\mathcal{F} \mid \boldsymbol{\Theta})] &\approx  \mathbf{E}_{q(\tau)} [\ln p(\mathcal{F} \mid \{\hat{\mathcal{G}}^{(k)}\}_{k \in [3]}, \hat{\mathcal{C}},\hat{\mathcal{S}},\tau )] \notag\\
  & = \sum_{i_1=1}^{I_1} \sum_{i_2=1}^{I_2} \sum_{i_3 =1}^{I_3}  \frac{1}{2} \ln \left\langle \tau \right\rangle - \frac{\left\langle\tau \right\rangle}{2} \Bigl( \mathcal{F}_{i_1 i_2 i_3} -   \mathrm{TW} [\{\hat{\mathcal{G}}^{(k)}\}_{k \in [3]}; \hat{\mathcal{C}}]_{i_1 i_2 i_3} - \hat{\mathcal{S}}_{i_1 i_2 i_3}\Bigl)^{2}.
\end{align}
Consequently, the optimization problem in (\ref{eq.5}) can be further approximated as
\begin{equation}
  \min_{q(\boldsymbol{\Theta})} \mathrm{KL}[q(\boldsymbol{\Theta}) || p(\boldsymbol{\Theta})] -  \mathbf{E}_{q(\tau)} [\ln p(\mathcal{F} \mid \{\hat{\mathcal{G}}^{(k)}\}_{k \in [3]}, \hat{\mathcal{C}},\hat{\mathcal{S}},\tau)], \label{eq.14}
\end{equation}
and we denote the objective function in this optimization problem as $L_{\text{VB}}$ for clarity.

Next, we proceed to calculate the first term. According to the variational posterior (\ref{VB pos}) and the prior distribution of the variables (see Appendix \ref{probabilistic Modeling} for more details), the first term in (\ref{KL expand}) can be expressed as:
\begin{align}
& \mathrm{KL}[q(\boldsymbol{\Theta}) || p(\boldsymbol{\Theta})] \notag \\
&=\mathrm{KL}\left[\prod_{k \in [3]} q\left(\mathcal{G}^{(k)}\right) q\left(\mathcal{C}\right) \prod_{k \in [3]} q\left(\boldsymbol{\lambda}_R^{(k)}\right)  \prod_{k \in [3]} q\left(\boldsymbol{\lambda}_L^{(k)}\right) || \right.\notag \\
&\qquad\quad\prod_{k \in [3]} p\left(\mathcal{G}^{(k)}\mid \boldsymbol{\lambda}^{(k)}_{R}, \boldsymbol{\lambda}^{(k+1)}_{R}, \boldsymbol{\lambda}^{(k)}_{L}\right)p(\mathcal{C} \mid \{\boldsymbol{\lambda}_{L}^{(k)}\}_{k \in [3]}) x\left. \prod_{k \in [3]} p(\boldsymbol{\lambda}_R^{(k)}) \prod_{k \in [3]} p(\boldsymbol{\lambda}_L^{(k)})\right] \notag \\
&\quad+ \mathrm{KL}[q(\tau) || p(\tau)] + \mathrm{KL}[q(\mathcal{S})q(\boldsymbol{\beta}) || p(\mathcal{S} \mid \boldsymbol{\beta})p(\boldsymbol{\beta})]. \label{KL solve}
\end{align}

When given the network outputs $[\langle \mathcal{G}^{(k)}\rangle, \sigma_{\mathcal{G}^{(k)}}^{2} ]$, $[\langle \mathcal{C}\rangle, \sigma_{\mathcal{C}}^{2}]$, and $[\langle \mathcal{S}\rangle, \sigma_{\mathcal{S}}^{2}]$, as in (\ref{output G})-(\ref{output S}), the KL divergence in (\ref{KL solve}) can be minimized in closed form with respect to $\boldsymbol{\lambda}^{(k)}_{R}$, $\boldsymbol{\lambda}^{(k)}_{L}$, $\tau$, $\boldsymbol{\beta}$ via VI theory, and the expressions can be found in (\ref{mean lambda_r}), (\ref{mean lambda_l}), (\ref{mean_tau}), and (\ref{mean_beta}) in Appendix \ref{close form}, respectively.

Then, the overall Variational Bayes (VB) loss function $L_{\text{VB}}$ in (\ref{eq.14}) for the Bayesian tensor decomposition module can be further derived as (see Appendix \ref{deri loss} for detailed derivations)
\begin{equation}
  L_{\text{VB}} = L_{\mathcal{F}} + L_{\langle \mathcal{G}^{(k)}\rangle} + L_{\sigma_{g^{k}}^{2}} + L_{\langle  \mathcal{C}  \rangle} + L_{\sigma_{C}^{2}} + L_{\langle \mathcal{S}\rangle} + L_{\sigma_{S}^{2}}. \label{loss VB}
\end{equation}
For clarity, the expressions and interpretations for items in (\ref{loss VB}) are summarized in Table \ref{tab:loss VB}.

Different from traditional approaches that add regularization terms to the loss function to prevent overfitting, with the Bayesian method, the loss terms in (\ref{loss VB}) are automatically traded off.
\begin{figure}[tbp]  
  \centering  
  \includegraphics[width=8cm]{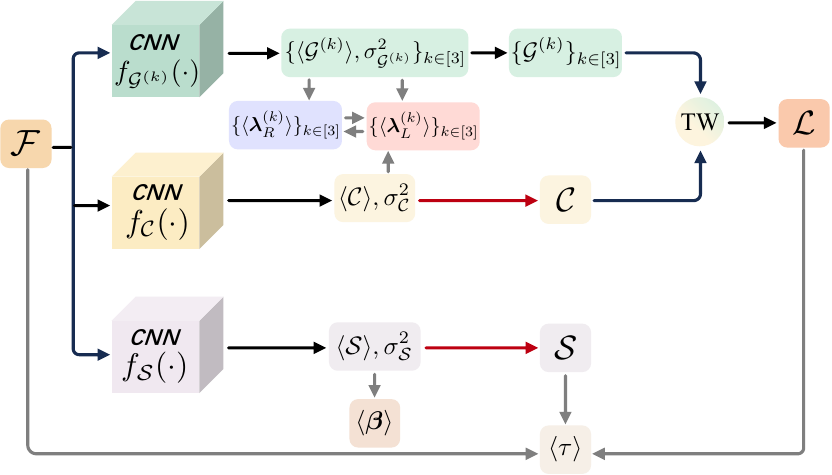} 
  \caption{The architecture of the Bayesian tensor decomposition module. Given a feature $\mathcal{F}$, we first use networks to estimate the mean $\langle \mathcal{G}_{t}^{(k)}\rangle$ and variance $\sigma_{\mathcal{G}_{t}^{(k)}}^{2} $ of the TW ring factors $\mathcal{G}_{t}^{(k)}$, $k\in[3]$, the mean $\langle \mathcal{C}_{t}\rangle$ and variance $\sigma_{\mathcal{C}_{t}}^{2}$ of the TW core factor $\mathcal{C}_{t}$, and the mean  $\langle \mathcal{S}_{t}\rangle$ and variance $\sigma_{\mathcal{S}_{t}}^{2}$ of the sparse component $\mathcal{S}_{t}$. Then, we apply the reparameterization trick to draw samples from these distributions. Subsequently, we compute the low-rank component $\mathcal{L}$ w.r.t the sampled ring factors $\{\mathcal{G}_{t}^{(k)}\}_{k \in [3]} $ and the sampled core factor $\mathcal{C}_{t}$ via the TW operation (\ref{TW model}). Meanwhile, we can compute $\langle \boldsymbol{\lambda}_{R}^{(k)} \rangle$, $\langle \boldsymbol{\lambda}_{L}^{(k)} \rangle$, $\langle\tau\rangle$ and $\langle \beta\rangle$, in closed-form via (\ref{mean lambda_r}), (\ref{mean lambda_l}), (\ref{mean_tau}), and (\ref{mean_beta}) in Appendix \ref{close form}, respectively. Then, the VB loss function (\ref{loss VB}) can be computed, so as to train the neural networks $f_{\mathcal{G}^{(k)}}(\cdot)$,$f_{\mathcal{C}}(\cdot)$, and $f_{\mathcal{S}}(\cdot)$. This training process can be jointly performed with that of the reconstruction-based AE (see Section~\ref{training strategy}).}  
  \Description{A flow diagram of the Bayesian tensor decomposition module, where neural networks estimate tensor factor distributions, samples are drawn by reparameterization, a low-rank component is computed by the tensor wheel operation, and a variational Bayes loss is formed.}
  \label{Bayes_decomp}
\end{figure}

By integrating our designed networks (\ref{output G})-(\ref{output S}), the adopted reparameterization trick (\ref{repa1})-(\ref{repa3}), and the closed-form solutions (\ref{mean lambda_r}), (\ref{mean lambda_l}), (\ref{mean_tau}), and (\ref{mean_beta}) in Appendix \ref{close form}, the overall Bayesian tensor decomposition module is illustrated in Fig.~\ref{Bayes_decomp}. Given a feature $\mathcal{F}$, we use networks to estimate the mean $\langle \mathcal{G}_{t}^{(k)}\rangle$ and variance $\sigma_{\mathcal{G}_{t}^{(k)}}^{2} $ of the TW ring factors $\mathcal{G}_{t}^{(k)} $, $k\in[3]$, the mean $\langle \mathcal{C}_{t}\rangle$ and variance $\sigma_{\mathcal{C}_{t}}^{2}$ of the TW core factor $\mathcal{C}_{t}$, and the mean  $\langle \mathcal{S}_{t}\rangle$ and variance $\sigma_{\mathcal{S}_{t}}^{2}$ of the sparse component $\mathcal{S}_{t}$. Then, we apply the reparameterization trick to draw samples from these distributions. Subsequently, we compute the low-rank component $\mathcal{L}$ w.r.t the sampled ring factors $\{\mathcal{G}_{t}^{(k)}\}_{k \in [3]} $ and the sampled core factor $\mathcal{C}_{t}$ via the TW operation (\ref{TW model}). Besides, $\langle \boldsymbol{\lambda}_{R}^{(k)} \rangle$,$\langle \boldsymbol{\lambda}_{L}^{(k)} \rangle$, $\langle\tau\rangle$, and $\langle \beta\rangle$ can be computed in closed-form via (\ref{mean lambda_r}), (\ref{mean lambda_l}), (\ref{mean_tau}), and (\ref{mean_beta}) in Appendix \ref{close form}, respectively. Then, the VB loss function (\ref{loss VB}) can be computed, so as to train the neural networks $f_{\mathcal{G}^{(k)}}(\cdot)$, $f_{\mathcal{C}}(\cdot)$, and $f_{\mathcal{S}}(\cdot)$. This training process can be jointly performed with that of the reconstruction-based AE (see Section~\ref{training strategy}).

So far, we have completed the design of the Bayesian tensor decomposition module. In the following, we introduce the mechanism by which the Bayesian tensor decomposition module mitigates the over-generalization problem of AE.

\subsubsection{The Mechanism for Mitigating the Over-generalization Problem}
We interpret the mechanism of the Bayesian tensor decomposition module in mitigating the over-generalization problem of AE in this subsection.

\begin{figure}[tbp]  
  \centering  
  \includegraphics[width=8cm]{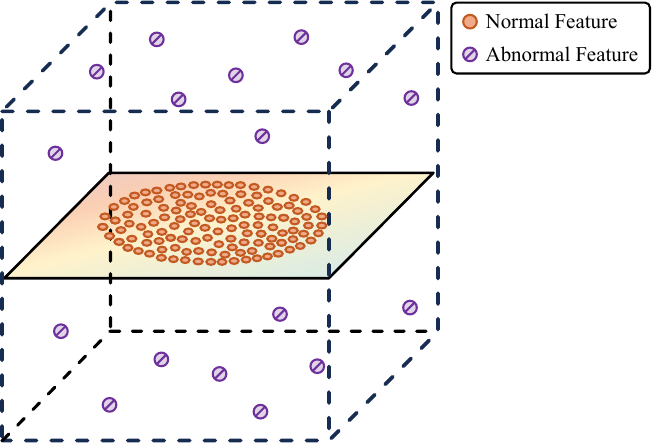} 
  \caption{The schematic feature distribution of normal and abnormal samples after applying the low-rank constraint. A key function of TW decomposition is its ability to perform dimensionality reduction for each tensor order, which compresses the normal features into a ``smaller'' space. For illustrative purposes, the figure presents a simplified scenario where the tensor order is 1. In this scenario, 3-dimensional normal features are compressed into a 2-dimensional plane. Abnormal features, which are typically non-low-rank (or sparse), do not conform to this low-rank subspace and thus deviate from the normal features.}  
  \Description{A schematic three-dimensional feature space in which normal samples lie near a low-rank two-dimensional plane while abnormal samples deviate from that plane.}
  \label{low_rank}
\end{figure}

Since the model is trained only on normal samples and the loss function in (\ref{loss VB}) encourages the component $\mathcal{S}$ in (\ref{decomp}) to be as sparse as possible during training, after training we have $\mathcal{F} \approx \mathcal{L}$ for normal data. In this manner, we impose a low-rank constraint on the features. This constraint compresses the normal features into a ``smaller'' space,  thereby effectively constraining the reconstruction capability of the AE. To illustrate this effect, the schematic feature distribution after applying the low-rank constraint is depicted in Fig.~\ref{low_rank}.

\begin{figure}[t]
	\centering
	\includegraphics[width=\textwidth]{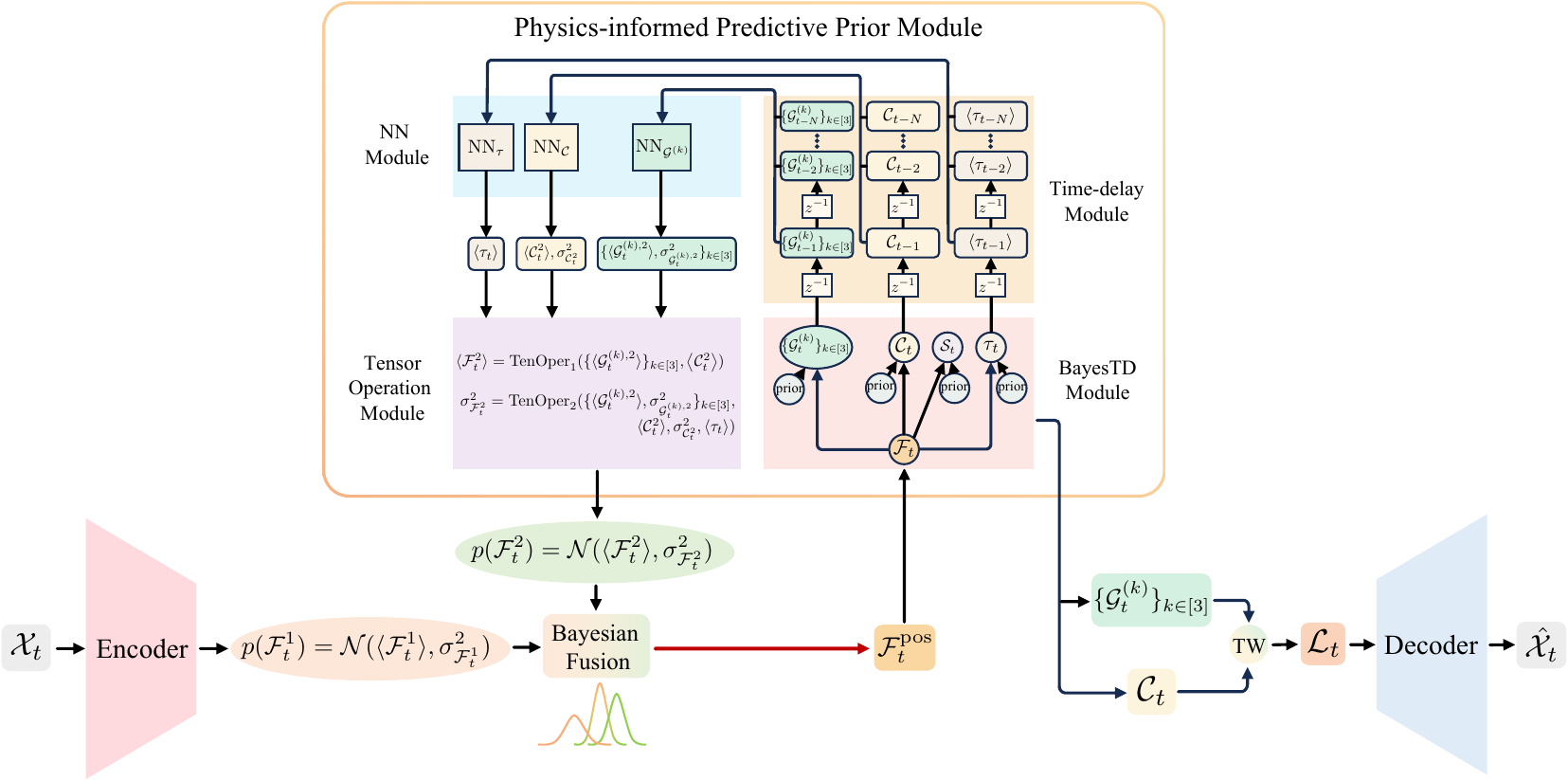}
	\caption{The architecture of PPPTAE, which consists of an encoder, a decoder, and a physics-informed predictive prior. For an observation $\mathcal{X}_{t}$ at a specific timestamp $t$, the overall forward process begins with the encoder, which infers the distribution $p( \mathcal{F}_t| \mathcal{X}_t)$ as defined in (\ref{likelihood}). Meanwhile, the physics-informed predictive prior predicts the distribution $p(\mathcal{F}_t | \mathcal{F}_{(t-M):(t-1)})$ as defined in (\ref{prior}). These two outputs are then fused using the Bayesian fusion approach detailed in Section \ref{Bayes fusion} to produce $p(\mathcal{F}_t | \mathcal{X}_t, \mathcal{F}_{(t-M):(t-1)})$. Following this, feature samples are approximately drawn from $p(\mathcal{F}_t | \mathcal{X}_t, \mathcal{F}_{(t-M):(t-1)})$ (as presented in Section \ref{training strategy}). These samples are subsequently processed by the Bayesian Tensor Decomposition (TD) module to estimate the TW ring factors $\{\mathcal{G}_{t}^{(k)}\}_{k \in [3]} $ and the TW core factor $\mathcal{C}_{t}$. Finally, the low-rank component $\mathcal{L}_{t}$ is calculated through the TW operation and then passed to the decoder for reconstruction.}
	\Description{A block diagram of PPPTAE showing the encoder, physics-informed predictive prior, Bayesian fusion, feature sampling, Bayesian tensor decomposition, tensor wheel operation, and decoder used to reconstruct the current observation.}
	\label{PPPTAE}
\end{figure}
Alternatively, from the perspective of coding length in information theory, we can also intuitively interpret the role of rank in the low-rank constraint. We know that normal data tends to be regular and thus easy to describe, allowing for short coding lengths. In contrast, anomalies are diverse and difficult to describe, leading to longer coding lengths. The rank in the low-rank constraint can be regarded as the coding length, and the role of the Bayesian tensor decomposition module is analogous to limiting the coding length of features, so as to mitigate the over-generalization problem of AE.

In the following, we present the overall design of the physics-informed predictive prior.

\subsubsection{Overall Design of the Physics-informed Predictive Prior}
The objective of the predictive prior is to predict the mean $\langle \mathcal{F}_{t} \rangle$ and variance $\sigma^{2}_{\mathcal{F}_{t}} $ of the feature $\mathcal{F}_{t}$ at timestamp $t$ as shown in (\ref{pp mean}) and (\ref{pp var}), respectively. According to the tensor decomposition framework (\ref{decomp}) and $\mathcal{L},\mathcal{S},\mathcal{E}$ are independent (see Appendix \ref{probabilistic Modeling} for details), we have
\begin{align}
\langle  \mathcal{F}_{t} \rangle &= \mathbf{E}[\mathcal{L}_{t}] + \mathbf{E}[\mathcal{S}_{t}] + \mathbf{E}[\mathcal{E}_{t}] \label{mean decomp}\\ 
\sigma^{2}_{\mathcal{F}_{t}}  &= \operatorname{var}[\mathcal{L}_{t}] + \operatorname{var}[\mathcal{S}_{t}] + \operatorname{var}[\mathcal{E}_{t}]. \label{var decomp}
\end{align}
Since the model is trained only on normal samples and the normal features are captured by the low-rank components $\mathcal{L}_{t}$, the sparse components $\mathcal{S}_{t}$ can be neglected. Moreover, the mean of the noise component $\mathcal{E}_{t}$ is zero and the variance is denoted by $\langle \tau_{t}\rangle$. Consequently, (\ref{mean decomp}) and (\ref{var decomp}) can be further rewritten as
\begin{align}
\langle  \mathcal{F}_{t} \rangle &= \mathbf{E}[\mathcal{L}_{t}] \label{mean decomp2} \\ 
\sigma^{2}_{\mathcal{F}_{t}} &= \operatorname{var}[\mathcal{L}_{t}] + \langle \tau_{t}\rangle. \label{var decomp2}
\end{align}
Furthermore, as the TW model is employed to characterize the low-rank structure of the feature tensor $\mathcal{F}$, as shown in (\ref{TW model}), and the variational posterior distributions of both the TW ring factors ${\mathcal{G}^{(k)}}_{k \in [3]}$ and the TW core factor $\mathcal{C}$ follow Gaussian distributions (see Appendix \ref{optimal} for details), (\ref{mean decomp2}) and (\ref{var decomp2}) can be further rewritten as
\begin{align}
\langle  \mathcal{F}_{t} \rangle  &= \text{TenOper}_1( \{\langle \mathcal{G}_{t}^{(k)} \rangle \}_{k \in [3]}, \langle \mathcal{C}_{t} \rangle) \\
\sigma^{2}_{\mathcal{F}_{t}}  &= \text{TenOper}_2( \{ \langle \mathcal{G}_{t}^{(k)} \rangle , \sigma_{\mathcal{G}_{t}^{(k)}}^{2} \}_{k \in [3]}, \langle \mathcal{C}_{t} \rangle, \sigma_{\mathcal{C}_{t}}^{2},  \langle  \tau_{t}\rangle ), 
\end{align}
where $\text{TenOper}_1(\cdot)$ and $\text{TenOper}_2(\cdot)$ are tensor operations, and the specific forms can be found in the Appendix.

To estimate the mean $\langle \mathcal{G}_{t}^{(k)}\rangle$ and variance $\sigma_{\mathcal{G}_{t}^{(k)}}^{2} $ of the TW ring factors $\mathcal{G}_{t}^{(k)} $ ($k \in [3]$), the mean $\langle \mathcal{C}_{t}\rangle$ and variance $\sigma_{\mathcal{C}_{t}}^{2}$ of the TW core factor $\mathcal{C}_{t}$, as well as the variance of the noise component $\langle \tau_{t}\rangle$ at timestamp $t$, we design predictive networks that leverage historical values for prediction, that is,
\begin{align}
[\langle \mathcal{G}_{t}^{(k)}\rangle, \sigma_{\mathcal{G}_{t}^{(k)}}^{2} ]&= \text{NN}_{\mathcal{G}^{(k)}}(\mathcal{G}_{t-1}^{(k)},\mathcal{G}_{t-2}^{(k)},\cdots,\mathcal{G}_{t-M}^{(k)})\\
[\langle \mathcal{C}_{t}\rangle, \sigma_{ \mathcal{C}_{t}}^{2}  ] &= \text{NN}_{\mathcal{C}}(\mathcal{C}_{t-1},\mathcal{C}_{t-2},\cdots,\mathcal{C}_{t-M})\\
\langle  \tau_{t}\rangle   &= \text{NN}_{\tau}(\langle  \tau_{t-1}\rangle ,\langle  \tau_{t-2}\rangle,\cdots, \langle  \tau_{t-M}\rangle ),
\end{align}
where $\text{NN}_{\mathcal{G}^{(k)}}, k\in [3], \text{NN}_{\mathcal{C}},$ and $\text{NN}_{\tau}$ are the predictive networks with convolutional layers.

With the physics-informed predictive prior integrated into the reconstruction-based TAE, the overall PPPTAE framework is illustrated in Fig. \ref{PPPTAE}. Given an observation $\mathcal{X}_{t}$ at timestamp $t$, the encoder first infers the distribution $p( \mathcal{F}_t| \mathcal{X}_t)$ as formulated in (\ref{likelihood}). Then, based on the outputs of both the physics-informed predictive prior and the encoder, the conditional distribution $p(\mathcal{F}_t | \mathcal{X}_t, \mathcal{F}_{(t-M):(t-1)})$ is obtained via the Bayesian fusion approach described in Section \ref{Bayes fusion}. Subsequently, samples are approximately drawn from $p(\mathcal{F}_t | \mathcal{X}_t, \mathcal{F}_{(t-M):(t-1)})$ (see Section \ref{training strategy} for details) and are fed into the Bayesian Tensor Decomposition (TD) module within the predictive prior to estimate the TW ring factors $\{\mathcal{G}_{t}^{(k)}\}_{k \in [3]} $ and the TW core factor $\mathcal{C}_{t}$. Finally, the low-rank component $\mathcal{L}_{t}$ is computed via the TW operation and is fed to the decoder for reconstruction.

\textbf{\textit{Remark:}} One may think to employ neural networks to directly predict the mean and variance of $\mathcal{F}$ using the historical features, as it may simplify the design of the predictive prior. However, since we expect $\mathcal{F}$ to possess the low-rank property so as to mitigate the over-generalization problem of AE, the corresponding low-rank structure is difficult to be directly learned by neural networks, and it is nontrivial for their outputs to directly have such a structure. In contrast, as shown in (\ref{TW model}), the TW model imposes no structural constraints on the ring factors $ \left\{\mathcal{G}^{(k)}\right\}_{k\in[3]}$ and the core factor $\mathcal{C}$. This means that, regardless of the specific forms of $ \left\{\mathcal{G}^{(k)}\right\}_{k\in[3]}$ and $\mathcal{C}$, the corresponding tensor obtained via the TW operation is always a low-TW-rank tensor. Therefore, our predictive networks are designed to utilize the decomposition results from the Bayesian tensor decomposition module for prediction.

\section{Training and Testing Strategies}

In this section, we introduce the strategies for training and testing PPPTAE.

\subsection{Training Strategy} \label{training strategy}
We only use normal data to train PPPTAE and all training data can be represented as a tensor with a size of $N_1 \times N_2 \times C \times T$, where $N_1$ and $N_2$ correspond to certain physical meanings (e.g., corresponding to the number of longitude and latitude points in weather data), $\mathcal{C}$ denotes the number of features, $T$ denotes the total number of timestamps.       

Since PPPTAE can utilize both the observations from the previous $M$ timestamps and the current observation, a natural idea to train PPPTAE is to scan the training dataset sequentially, from start to end, thereby completing a training epoch. However, this approach introduces insufficient randomness during training, potentially leading to the problem of catastrophic forgetting or overfitting in the model. Specifically, the model may learn the later parts of the training sequences well but forget the earlier ones, resulting in inferior performance.

To address this issue, we randomly sample the training data with a window length of $M+1$, resulting in a training sample sized $N_1 \times N_2 \times C \times (M+1)$. The first $M$ tensors in this training sample are fed into the encoder individually, and then we directly sample their features, which are used as the input for the physics-informed predictive prior. Next, the last tensor in this training sample is fed into the encoder. Based on the results of the physics-informed predictive prior and the encoder, we can obtain $p(\mathcal{F}_t | \mathcal{X}_t, \mathcal{F}_{(t-M):(t-1)})$ using the Bayesian fusion approach described in Section \ref{Bayes fusion}.

Subsequently, to simplify analysis and for computational tractability, we sample from $p(\mathcal{F}_t | \mathcal{X}_t, \\\mathcal{F}_{(t-M):(t-1)})$, i.e., $\mathcal{F}_t \sim p(\mathcal{F}_t | \mathcal{X}_t, \mathcal{F}_{(t-M):(t-1)})$, and then use the Bayesian tensor decomposition module described in Section \ref{Bayesian TD} to obtain the corresponding low-rank component $\mathcal{L}_t$. This low-rank component is then fed into the decoder to obtain the reconstruction $\hat{\mathcal{X}}=D(L(\mathcal{F}_{t}))$, where $L(\cdot)$ denotes the operation for estimating the low-rank component via the Bayesian tensor decomposition module, and $D(\cdot)$ denotes the reconstruction operation performed by the decoder. Based on the reconstruction $\hat{\mathcal{X}}=D(L(\mathcal{F}_{t}))$ and the corresponding original observation $\mathcal{X}$, the objective function is designed as:
\begin{align}
L_{\text{Recon}} &= \mathbf{E}_{\mathcal{F}_{t} \sim p(\mathcal{F}_t | \mathcal{X}_t, \mathcal{F}_{(t-M):(t-1)})}[\| \hat{\mathcal{X}}_{t} - \mathcal{X}_{t}\|_{F}^{2}] \notag \\
&=\mathbf{E}_{\mathcal{F}_{t} \sim p(\mathcal{F}_t | \mathcal{X}_t, \mathcal{F}_{(t-M):(t-1)})}[\| D(L(\mathcal{F}_{t}) - \mathcal{X}_{t}\|_{F}^{2}]. \label{loss1}
\end{align}

\begin{algorithm}[tbp]
  \caption{Training of PPPTAE} 
  \label{alg1}
  \begin{algorithmic}[1]
  \REQUIRE Training datasets.
  \ENSURE Parameters of PPPTAE.

  \FOR{$n=1:N_{\text{epoch}}$}
    \STATE Sample training data with window length $M+1$.
    \STATE Feed the first $M$ tensors into the encoder individually to obtain $p(\mathcal{F}_t\mid\mathcal{X}_t)$ (see (\ref{likelihood})).
    \STATE Sample from $p(\mathcal{F}_t\mid\mathcal{X}_t)$ to get features and use them as input to the physics-informed predictive prior to obtain $p(\mathcal{F}_t\mid\mathcal{F}_{(t-M):(t-1)})$ (see (\ref{prior})).
    \STATE Feed the last tensor into the encoder to obtain $p(\mathcal{F}_t\mid\mathcal{X}_t)$.
    \STATE Compute $\mu_{\text{fused}}$ and $\sigma_{\text{fused}}^2$ via (\ref{mu_fuse}) and (\ref{sigma_fuse}).
    \STATE Update the parameters of PPPTAE via (\ref{total loss}).
  \ENDFOR
  \RETURN Parameters of PPPTAE.

  \end{algorithmic}
\end{algorithm}

\begin{algorithm}[tbp]
  \caption{Testing of PPPTAE} 
  \label{alg2}
  \begin{algorithmic}[1]
  \REQUIRE Testing datasets.
  \ENSURE Anomaly scores of the testing datasets.
  
  \FOR{$t=1:T_{\text{test}}$}
    \IF{\textit{Cold-start} ($t\le M$)}
      \STATE Estimate mean $\langle\mathcal{F}_t^{1}\rangle$ of $p(\mathcal{F}_t\mid\mathcal{X}_t)$ (see (\ref{likelihood})) via the encoder.
      \STATE Obtain low-rank component $\mathcal{L}_t$ of $\langle\mathcal{F}_t^{1}\rangle$ via the BayesTD module.
    \ELSE
      \STATE Compute $\mu_{\text{fused}}$ via (\ref{mu_fuse}).
      \STATE Obtain low-rank component $\mathcal{L}_t$ of $\mu_{\text{fused}}$ via the BayesTD module.
    \ENDIF
    \STATE Cache $\mathcal{L}_t$ in the time-delay module.
    \STATE Feed $\mathcal{L}_t$ to the decoder to obtain reconstruction $\hat{\mathcal{X}}_t$.
    \STATE Compute the anomaly score via (\ref{ano score}).	
  \ENDFOR
  
  \RETURN Anomaly scores.
  \end{algorithmic}
\end{algorithm}

To sample from $p(\mathcal{F}_t | \mathcal{X}_t, \mathcal{F}_{(t-M):(t-1)})$, a natural idea is to try to employ the reparameterization trick similar to that used in (\ref{repa1})-(\ref{repa3}) for sampling. However, due to the presence of the constant term $\alpha = p(\mathcal{X}_t)/ p(\mathcal{X}_t | \mathcal{F}_{(t-M):(t-1)})$ and the probabilistic distribution $p(\mathcal{F}_{t})$ in the denominator as shown in (\ref{posterior}), the reparameterization trick is not feasible. To solve this problem, inspired by the importance resampling method \cite{impor1,impor2,impor3}, we choose $q(\mathcal{F}_t) = \mathcal{N}(\mathcal{F}_t; \mu_{\text{fused}}, \sigma_{\text{fused}}^2)$ as the proposal distribution, where $\mu_{\text{fused}}$ and $\sigma_{\text{fused}}^2$ can be computed via (\ref{mu_fuse}) and (\ref{sigma_fuse}), respectively. Next, we sample directly from the proposal distribution, and the loss function (\ref{loss1}) can be equivalently transformed into
\begin{align}
L_{\text{Recon}} =&\mathbf{E}_{\mathcal{F}_{t} \sim q(\mathcal{F}_t )}\left[\frac{p(\mathcal{F}_t | \mathcal{X}_t, \mathcal{F}_{(t-M):(t-1)})}{q(\mathcal{F}_{t})}\| D(L(\mathcal{F}_{t}) - \mathcal{X}_{t}\|_{F}^{2}\right] \notag \\
\approx &  \sum_{n=1}^{N}  w_{n}\| D(L(\mathcal{F}_{t})) - \mathcal{X}_{t}\|_{F}^{2},  \label{loss2} 
\end{align}
where
\begin{align}
w_{n} = \frac{\tilde{w}_{n}}{\sum_{n'}\tilde{w}_{n'}}, \quad \tilde{w}_{n} &= \frac{p(\mathcal{F}_t^{(n)} | \mathcal{X}_t, \mathcal{F}_{(t-M):(t-1)})}{q(\mathcal{F}_{t}^{(n)})}. \label{weight}
\end{align}
When substituting (\ref{posterior}) into (\ref{weight}), we have
\begin{equation}
  w_{n} = \frac{\frac{1}{p(\mathcal{F}_{t}^{(n)})}}{\sum_{n'}\frac{1}{p(\mathcal{F}_{t}^{(n')})}},
\end{equation}
which means that samples located in sparse regions receive greater weight in the loss function (\ref{loss2}). This property enables the model to reconstruct the normal samples well in sparse regions.

To estimate $p(\mathcal{F}_{t})$, (\ref{int Ft}) shows that this probability density, in theory, requires integrating over all possible inputs $\mathcal{X}_{t}$, which is computationally intractable in practice. Therefore, we adopt an approximate approach by using the features $\mathcal{F}_{(t-M):(t-1)}$ from the previous $M$ timestamps within a training mini-batch as reference points. We then estimate the probability density value $p(\mathcal{F}_t)$ of the feature $\mathcal{F}_t$ using KDE with the Gaussian kernel. 

By integrating the loss function $L_{\text{VB}}$ of the Bayesian tensor decomposition module, as defined in (\ref{loss VB}), the overall training objective of PPPTAE is formulated as
\begin{equation}
  L = L_{\text{Recon}} + \lambda L_{\text{VB}}, \label{total loss}
\end{equation} 
where $\lambda$ serves as a trade-off hyperparameter. The training algorithm of PPPTAE is shown in Algorithm \ref{alg1}.

\subsection{Testing Strategy}


During the testing stage, we scan the testing dataset sequentially, from start to end, so as to obtain the testing results. At the initial cold-start stage, where the number of samples already tested is less than $M$, the predictive prior does not work. Instead, we directly feed the observation into the encoder, and the mean of the feature is further processed by the Bayesian tensor decomposition module for reconstruction. In this process, the time-delay module would cache the results of the Bayesian tensor decomposition module.

Once the cache size of the time-delay module in the predictive prior reaches $M$, the predictive prior begins to work. At this time, we fuse the results of the predictive prior and the encoder via (\ref{mu_fuse}), and the mean is then further processed by the Bayesian tensor decomposition module for reconstruction. In this process, the time-delay module would append the newest results of the Bayesian tensor decomposition module, and pop the oldest results to maintain a constant perception length.

For the anomaly score, we use the deviation between the reconstruction $\hat{\mathcal{X}}_{t}$ and the observation $\mathcal{X}_{t}$, and quantify this deviation using the squared Frobenius norm, i.e.,
\begin{equation}
  Score_{t} = \|\mathcal{X}_{t} - \hat{\mathcal{X}}_{t} \|^{2}_{F}. \label{ano score}
\end{equation}
The testing algorithm of PPPTAE is shown in Algorithm \ref{alg2}.

\section{Experiments}

In this section, we conduct experiments to demonstrate the effectiveness of PPPTAE.

\subsection{Datasets}

We conduct experiments on three representative multi-dimensional time series anomaly detection datasets, i.e, the Freeway dataset \cite{Freeway}, the HR-Extreme dataset \cite{Extreme}, and the Milan \cite{Milan} dataset, to evaluate the proposed PPPTAE.

The Freeway dataset is a large-scale lane-level freeway traffic dataset. This dataset consists of a month of weekday traffic data recorded by 49 radar detection sensors. The traffic speed, occupancy, and volume data are collected for the 4 interstate lanes every 30 seconds. This dataset can be represented by a multi-dimensional time series sized $49 \times 4 \times 3 \times T$. Abnormal events, such as crashes, are recorded in this dataset.

The HR-Extreme dataset is a high-resolution weather dataset. This dataset consists of a year of weather feature maps sized $320 \times 320$ in the U.S., with each entry corresponding to a 3km by 3km area. 69 physical variables, such as temperature and humidity, are collected as the channels of the feature maps every hour. This dataset can be represented by a multi-dimensional time series sized $320 \times 320 \times 69 \times T$. Seventeen abnormal extreme weather events, such as hurricanes, tornadoes, and severe storms, are recorded in this dataset.

The Milan dataset records the telecommunication activities of the city of Milan from November 1st, 2013, to January 1st, 2014. The data are aggregated in a grid with $100 \times 100$ geographical square cells, with each cell corresponding to a 235m by 235m area. 5 activities including received Short Message Service (SMS), sent SMS, incoming call, outgoing call, and internet connection, are recorded with a temporal aggregation of time slots of one hour. This dataset can be represented by a multi-dimensional time series sized $100 \times 100 \times 5 \times T$. Abnormal patterns, such as the increase in network activities and the number of calls on New Year's Eve, are recorded in this dataset.

\begin{table}[]
\centering
\caption{The Architecture of the Encoder and Decoder of PPPTAE}
\label{tab:AE}
\resizebox{0.7\columnwidth}{!}{%
\begin{tabular}{@{}cccccc@{}}
\toprule
\multicolumn{3}{c|}{Encoder} &
  \multicolumn{3}{c}{Decoder} \\ \midrule
\multicolumn{1}{c|}{Stage} &
  Layer &
  \multicolumn{1}{c|}{Output Size} &
  \multicolumn{1}{c|}{Stage} &
  Layer &
  Output Size \\ \midrule
\multicolumn{1}{c|}{Input} &
  - &
  \multicolumn{1}{c|}{$C \times N_1 \times   N_2$} &
  \multicolumn{1}{c|}{Input} &
  Sampled $\mathcal{F}$ &
  $256 \times N_1'' \times   N_2''$ \\ \midrule
\multicolumn{1}{c|}{\multirow{2}{*}{Stage 1}} &
  $\mathcal{F}_{conv}(C,64)$ &
  \multicolumn{1}{c|}{$64\times N_1 \times N_2$} &
  \multicolumn{1}{c|}{\multirow{3}{*}{Stage   1}} &
  $\mathcal{F}_{conv}(256,256)$ &
  $256 \times N_1'' \times   N_2''$ \\ \cmidrule(lr){2-3} \cmidrule(l){5-6} 
\multicolumn{1}{c|}{} &
  MaxPool ($\psi(\cdot)$) &
  \multicolumn{1}{c|}{$64\times N_1' \times   N_2'$} &
  \multicolumn{1}{c|}{} &
  Bilinear($N_1',N_2'$) &
  $256 \times N_1' \times   N_2'$ \\ \cmidrule(r){1-3} \cmidrule(l){5-6} 
\multicolumn{1}{c|}{\multirow{2}{*}{Stage 2}} &
  $\mathcal{F}_{conv}(64,128)$ &
  \multicolumn{1}{c|}{$128 \times N_1' \times   N_2'$} &
  \multicolumn{1}{c|}{} &
  Conv $3 \times3$, ReLU &
  $128 \times N_1' \times   N_2'$ \\ \cmidrule(l){2-6} 
\multicolumn{1}{c|}{} &
  MaxPool ($\psi(\cdot)$) &
  \multicolumn{1}{c|}{$128 \times N_1'' \times   N_2''$} &
  \multicolumn{1}{c|}{\multirow{3}{*}{Stage   2}} &
  $\mathcal{F}_{conv}(128,128)$ &
  $128 \times N_1' \times   N_2'$ \\ \cmidrule(r){1-3} \cmidrule(l){5-6} 
\multicolumn{1}{c|}{Stage 3} &
  $\mathcal{F}_{conv}(128,256)$ &
  \multicolumn{1}{c|}{$256 \times N_1'' \times   N_2''$} &
  \multicolumn{1}{c|}{} &
  Bilinear($N_1,N_2$) &
  $128 \times N_1 \times   N_2$ \\ \cmidrule(r){1-3} \cmidrule(l){5-6} 
\multicolumn{1}{c|}{\multirow{2}{*}{Latent}} &
  Conv $3 \times3$, BN &
  \multicolumn{1}{c|}{$256 \times N_1'' \times   N_2''$} &
  \multicolumn{1}{c|}{} &
  Conv $3 \times 3$, ReLU &
  $64 \times N_1 \times   N_2$ \\ \cmidrule(l){2-6} 
\multicolumn{1}{c|}{} &
  Conv $3 \times3$, BN &
  \multicolumn{1}{c|}{$256 \times N_1'' \times   N_2''$} &
  \multicolumn{1}{c|}{\multirow{3}{*}{Output}} &
  Conv $3 \times3$, ReLU &
  $16 \times N_1 \times   N_2$ \\ \cmidrule(r){1-3} \cmidrule(l){5-6} 
\multicolumn{1}{l}{} &
  \multicolumn{1}{l}{} &
  \multicolumn{1}{l|}{} &
  \multicolumn{1}{c|}{} &
  Conv $3 \times3$, ReLU &
  $16 \times N_1 \times   N_2$ \\ \cmidrule(l){5-6} 
\multicolumn{1}{l}{} &
  \multicolumn{1}{l}{} &
  \multicolumn{1}{l|}{} &
  \multicolumn{1}{c|}{} &
  Conv $3 \times3$, Sigmoid &
  $C \times N_1   \times N_2$ \\ \midrule
\multicolumn{6}{l}{Note: $N_1' =\psi(N_1)$, $N_2' =\psi(N_2)$ and $N_1'' =\psi(N_1')$, $N_2'' =\psi(N_2')$.}
\end{tabular}%
}
\end{table}

\begin{table}[]
\centering
\caption{The Architecture of the Networks in the Bayesian Tensor Decomposition Module}
\label{tab:BayesNN}
\resizebox{0.7\columnwidth}{!}{%
\begin{tabular}{@{}cc|cc@{}}
\toprule
\multicolumn{2}{c|}{$f_{\mathcal{G}^{(1)}}$}                         & \multicolumn{2}{c}{$f_{\mathcal{G}^{(2)}}$}                        \\ \midrule
Layer                          & Output Size                         & Layer                        & Output Size                         \\ \midrule
Trilinear($8,8,256$)           & $1 \times 8\times 8   \times 256$   & Trilinear($8,8,N_1''$)       & $1 \times 8\times 8   \times N_1''$ \\ \midrule
Conv $3 \times 3 \times 3$,   BN, LReLU & $4 \times 8\times 8   \times 256$   & Conv $3 \times 3 \times 3$, BN, LReLU & $4 \times 8\times 8   \times N_1''$  \\ \midrule
Conv $3 \times 3 \times 3$, BN          & $16 \times 8\times 8   \times 256$  & Conv $3 \times 3 \times 3$, BN        & $16 \times 8\times 8   \times N_1''$ \\ \bottomrule
\multicolumn{2}{c|}{$f_{\mathcal{G}^{(3)}}$}                         & \multicolumn{2}{c}{$f_{\mathcal{C}}$}                              \\ \midrule
Trilinear($8,8,N_2''$)         & $1 \times 8\times 8   \times N_2''$ & Bilinear($8,8$)              & $256 \times 8\times 8'$             \\ \midrule
Conv $3 \times 3 \times 3$,   BN, LReLU & $4 \times 8\times 8   \times N_2''$ & Conv $3 \times 3$, BN, LReLU          & $64\times 8\times 8'$                \\ \midrule
Conv $3 \times 3 \times 3$, BN & $16 \times 8\times 8 \times N_2''$  & Conv $3 \times 3$, BN        & $16\times 8\times 8'$               \\ \bottomrule
\multicolumn{1}{l}{}           & \multicolumn{1}{l|}{}               & \multicolumn{2}{c}{$f_{\mathcal{S}}$}                              \\ \cmidrule(l){3-4} 
\multicolumn{1}{l}{}           & \multicolumn{1}{l|}{}               & Conv $3 \times 3$, BN, LReLU & $256 \times N_1''\times   N_2''$    \\ \cmidrule(l){3-4} 
\multicolumn{1}{l}{}           & \multicolumn{1}{l|}{}               & Conv $3 \times 3 $, BN       & $512 \times N_1''\times   N_2''$    \\ \bottomrule
\end{tabular}%
}
\end{table}

\begin{table}[]
\centering
\caption{The Architecture of the Networks in the Neural Network (NN) Module}
\label{tab:NN module}
\resizebox{0.7\columnwidth}{!}{%
\begin{tabular}{@{}cc|cc@{}}
\toprule
\multicolumn{2}{c|}{$\text{NN}_{\mathcal{G}^{(1)}}$} & \multicolumn{2}{c}{$\text{NN}_{\mathcal{G}^{(2)}}$} \\ \midrule
Layer                    & Output Size               & Layer                    & Output Size              \\ \midrule
Conv $3 \times 3 \times 3$,   BN, LReLU & $16 \times 8\times 8   \times 256$   & Conv $3 \times 3 \times 3$, BN, LReLU & $16 \times 8\times 8   \times N_1''$ \\ \midrule
Conv $3 \times 3 \times 3$, BN          & $16 \times 8\times 8   \times 256$   & Conv $3 \times 3 \times 3$, BN        & $16 \times 8\times 8   \times N_1''$ \\ \bottomrule
\multicolumn{2}{c|}{$\text{NN}_{\mathcal{G}^{(3)}}$} & \multicolumn{2}{c}{$\text{NN}_{\mathcal{C}}$}       \\ \midrule
Conv $3 \times 3 \times 3$,   BN, LReLU & $16 \times 8\times 8   \times N_2''$ & Conv $3 \times 3$, BN, LReLU          & $16\times 8\times 8'$                \\ \midrule
Conv $3 \times 3 \times 3$, BN          & $16\times 8\times 8   \times N_2''$  & Conv $3 \times 3$, BN                 & $16\times 8\times 8'$                \\ \bottomrule
\multicolumn{1}{l}{}     & \multicolumn{1}{l|}{}     & \multicolumn{2}{c}{$\text{NN}_{\tau}$}              \\ \cmidrule(l){3-4} 
\multicolumn{1}{l}{}     & \multicolumn{1}{l|}{}     & Linear, ReLU             & $16 \times 1$            \\ \cmidrule(l){3-4} 
\multicolumn{1}{l}{}     & \multicolumn{1}{l|}{}     & Linear                   & $1\times 1$              \\ \bottomrule
\end{tabular}%
}
\end{table}

\begin{table}[t]
\centering
\caption{AUC of Time Series Anomaly Detection Methods}
\label{tab:SOTA}
\resizebox{\textwidth}{!}{%
\begin{tabular}{@{}cccccccc@{}}
\toprule
Method                                        & Venue      & Freeway       & HR-Extreme\_S1 & HR-Extreme\_S2 & HR-Extreme\_S3 & HR-Extreme\_Y & Milan         \\ \midrule
Reformer \cite{Reformer}     & ICLR-20    & 75.24\%       & 69.92\%        & 84.52\%        & 72.79\%        & 78.86\%       & 69.40\%       \\
AnoTrans \cite{AnoTrans}     & ICLR-21    & 72.71\%       & 71.51\%        & 72.81\%        & 76.46\%        & 71.75\%       & 68.43\%       \\
Autoformer \cite{Autoformer} & NeurIPS-21 & 69.15\%       & 72.66\%        & 70.70\%        & 70.23\%        & 70.53\%       & 83.87\%       \\
Informer \cite{Informer}     & AAAI-21    & 75.22\%       & 71.90\%        & 84.00\%        & 78.98\%        & 80.88\%       & 79.01\%       \\
TFAD \cite{TFAD}             & CIKM-22    & 71.74\%       & 70.19\%        & 70.21\%        & 80.44\%        & 71.48\%       & 80.36\%       \\
Fedformer \cite{Fedformer}   & ICML-22    & 69.44\%       & 78.80\%        & 74.04\%        & 84.79\%        & 78.72\%       & 74.05\%       \\
Pyraformer \cite{Pyraformer} &
  ICLR-22 &
  82.95\% &
  85.41\% &
  79.94\% &
  {\ul 90.90\%} &
  87.82\% &
  79.47\% \\
MICN \cite{MICN}             & ICLR-23    & 77.52\%       & 69.76\%        & 83.85\%        & 82.53\%        & 80.59\%       & 88.85\%       \\
DCdetector \cite{DCdetector} & KDD-23     & 84.55\%       & 70.59\%        & 72.35\%        & 81.77\%        & 76.78\%       & 86.80\%       \\
PatchTST \cite{PatchTST}     & ICLR-23    & 88.72\%       & 86.82\%        & 89.47\%        & 82.40\%        & 84.20\%       & {\ul 91.43\%} \\
Crossformer \cite{Crossformer} &
  ICLR-23 &
  71.55\% &
  {\ul 92.18\%} &
  82.85\% &
  87.22\% &
  86.84\% &
  88.69\% \\
NLinear \cite{NLinear}       & AAAI-23    & 76.94\%       & 82.11\%        & 79.37\%        & 88.57\%        & 83.15\%       & 61.63\%       \\
TimesNet \cite{TimesNet}     & ICLR-23    & 88.85\%       & 90.85\%        & {\ul 89.67\%}  & 89.24\%        & 89.64\%       & 83.85\%       \\
ModernTCN \cite{ModernTCN}   & ICLR-24    & {\ul 90.66\%} & OOM            & OOM            & OOM            & OOM           & OOM           \\
CATCH \cite{CATCH}           & ICLR-25    & 90.01\%       & OOM            & OOM            & OOM            & OOM           & OOM           \\
PPPTAE &
  This work &
  \textbf{95.39\%} &
  \textbf{98.82\%} &
  \textbf{92.21\%} &
  \textbf{95.28\%} &
  \textbf{94.65\%} &
  \textbf{98.90\%} \\ \bottomrule
\end{tabular}%
}
\vspace{2pt}
\begin{minipage}{\textwidth}
\footnotesize
The entities in bold denote the best results, and those with an underline are the second-best results. OOM is an abbreviation of out of memory.
\end{minipage}
\end{table}

\subsection{Implementation Details}

\subsubsection{Network Architecture}

We construct a 2D convolutional network backbone to preserve the multi-dimensional structure of the tensor data at each timestamp and the architecture of the encoder and decoder of PPPTAE is shown in Table I. In this table, for brevity, we use $\mathcal{F}_{conv}(C_{in},C_{out})$ to denote a basic convolutional block consisting of two stacked $3 \times 3$, stride 1, and padding 1 convolutional layers, each followed by a ReLU activation. Moreover, to handle arbitrary input size robustly, we employ an adaptive downsampling strategy. We define the spatial reduction function $\psi(d)$ for a dimension $d$ as:
\begin{equation}
  \psi(d) = \begin{cases} \max(1, \lfloor d/2 \rfloor), & \text{if } d > 2 \\ d, & \text{otherwise} \end{cases}
\end{equation}
The encoder utilizes adaptive max-pooling operations that map input spatial dimensions $N_1 \times N_2$ to $\psi(N_1) \times \psi(N_2)$. The decoder mirrors this process using adaptive bilinear upsampling to strictly recover the recorded spatial resolutions from the encoder.

We then construct the networks in the Bayesian tensor decomposition module (see Fig. \ref{Bayes_decomp}), and the architecture is shown in Table \ref{tab:BayesNN}. With such an architecture, the initial TW rank is set as $\mathbf{r=}$[R=[8,8,8,8], L=[8,8,8,8]]. Before training, we initialize the weights of $f_{\mathcal{S}}$ to zeros as $\mathcal{S}$ is a sparse tensor. Moreover, we construct the networks in the NN module (see Fig. \ref{PPPTAE}), and the architecture is shown in Table \ref{tab:NN module}.

\subsubsection{Paremeter Settings}

We use the Adam optimizer \cite{Adam} to train the model with a batch size of 64, and set the learning rate to 1e-4. We set the trade-off hyperparameter $\lambda$ in the training loss (\ref{total loss}) to 0.001, so that the two loss terms are of roughly the same magnitude. Without loss of generality, we set the number of previous timestamps $M=4$, the number of importance samples $N = 10$ in (\ref{loss2}), and the KDE bandwidth to 1. The training epochs are set to 400.

\subsection{Comparison with State-of-the-art}

We then show the comparison results between PPPTAE and State-Of-The-Art (SOTA) methods.

\subsubsection{Evaluation Metric}

We utilize the widely adopted Area Under the Curve (AUC) as the evaluation metric. By varying the threshold values for anomaly scores to generate the ROC curve, we compute the AUC. A higher AUC indicates superior anomaly detection performance.

\begin{table}[t]
\centering
\caption{Ablation Results of PPPTAE}
\label{tab:Ablation}
\resizebox{\textwidth}{!}{%
\begin{tabular}{@{}c|c|c|c|c|cccccc@{}}
\toprule
Index &
  AE\_r &
  AE\_p &
  BayesLRM &
  Predictive Prior &
  Freeway &
  HR-Extreme\_S1 &
  HR-Extreme\_S2 &
  HR-Extreme\_S3 &
  HR-Extreme\_Y &
  Milan \\ \midrule
1 &
  $\checkmark$ &
   &
   &
   &
  75.09\% &
  96.63\% &
  85.58\% &
  87.93\% &
  91.09\% &
  90.12\% \\
2 &
  $\checkmark$&
   &
  $\checkmark$ &
   &
  88.80\% &
  96.92\% &
  89.58\% &
  90.23\% &
  91.33\% &
  95.12\% \\
3 &
  $\checkmark$ &
   &
   &
  $\checkmark$ &
  86.20\% &
  97.41\% &
  86.35\% &
  91.45\% &
  92.03\% &
  94.28\% \\
4 &
   &
  $\checkmark$ &
   &
   &
  83.37\% &
  86.31\% &
  87.35\% &
  92.87\% &
  90.58\% &
  93.83\% \\
5* &
  $\checkmark$ &
   &
  $\checkmark$ &
  $\checkmark$ &
  \textbf{95.39\%} &
  \textbf{98.82\%} &
  \textbf{92.21\%} &
  \textbf{95.28\%} &
  \textbf{94.65\%} &
  \textbf{98.90\%} \\ \bottomrule
\end{tabular}
}
\vspace{2pt}
\begin{minipage}{\textwidth}
\footnotesize
The models AE\_r and AE\_p are the reconstruction-based and prediction-based TAE baselines of PPPTAE, respectively. BayesLRM is an abbreviation of Bayesian low-rank module. The index with * denotes the best-performing model and the best results are highlighted in bold.
\end{minipage}
\end{table}

\subsubsection{Experiments Settings}

We normalized the data of the datasets to $(0,1)$. For the HR-Extreme dataset, we further crop the spatial dimensions of the feature maps to $64 \times 64$ to ensure computational tractability. In addition, to demonstrate the anomaly detection performance of models in different periods of the year, we further divide the HR-Extreme dataset into three subsets covering January to April, May to August, and September to December, respectively denoted as HR-Extreme\_S1, HR-Extreme\_S2, and HR-Extreme\_S3. We denote the dataset covering the entire year as HR-Extreme\_Y.

\subsubsection{Results}

We compare the performance of PPPTAE with 15 SOTA methods on six benchmark datasets, and summarize the results in Table \ref{tab:SOTA}. We see that PPPTAE achieves the best AUC performance compared to the SOTA methods on each dataset. Note that the existing methods are conﬁned to multi-variate time series and cannot directly handle multi-dimensional time series. In this experiment, to apply these methods for multi-dimensional time series anomaly detection, we merge $N_1$, $N_2$, $C$. However, this operation inevitably breaks the intrinsic correlations and thus leads to performance degradation. In contrast, PPPTAE can directly handle multi-dimensional time series and better performance is achieved. These results demonstrate the effectiveness of PPPTAE for multi-dimensional time series anomaly detection.

\subsection{Ablation Studies}

In this subsection, we conduct ablation studies on the proposed PPPTAE. We analyze the effectiveness of the model components, including the Bayesian low-rank module and the predictive prior. We also compare PPPTAE with the baseline reconstruction-based and prediction-based TAE, which are denoted by AE\_r and AE\_p, respectively. The results are presented in Table \ref{tab:Ablation}. We see from Table \ref{tab:Ablation} that AE\_r with the Bayesian low-rank module outperforms AE\_r, as the Bayesian low-rank module imposes a low-rank constraint on the features, thereby compressing the features of normal samples into a ``smaller'' space, thus mitigating the over-generalization problem of AE\_r. Besides, AE\_r with the predictive prior performs better than  AE\_r, as the predictive prior can capture the temporal dependencies among features and provide historical information for AE\_r, so that the performance can be enhanced. With both the Bayesian low-rank module and the predictive prior, the performance is further enhanced. In addition, we note that AE\_r and AE\_p demonstrate complementary strengths, with neither consistently outperforming the other. This is because AE\_r is more effective at detecting abrupt anomalies, whereas AE\_p is better suited for identifying temporally correlated anomalies. Moreover, PPPTAE achieves better performance than both AE\_r and AE\_p, demonstrating  that PPPTAE can effectively integrate the prediction and reconstruction capabilities.

\subsection{Qualitative Studies}

In this subsection, we conduct qualitative studies on the proposed PPPTAE.

\begin{figure}[]  
  \centering  
  \includegraphics[width=7cm]{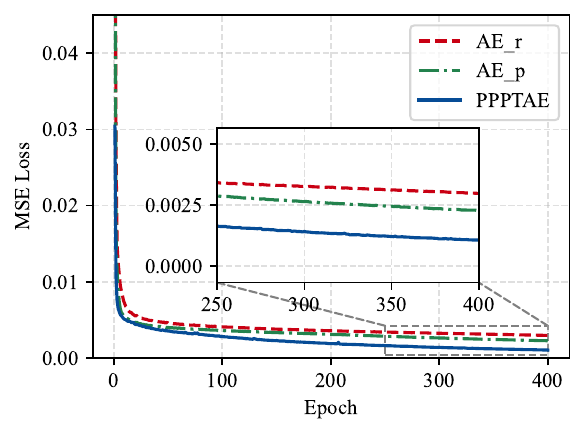} 
  \caption{The MSE loss of AE\_r, AE\_p, and PPPTAE on the Freeway dataset.}  
  \Description{A line chart comparing the mean squared error training losses of AE-r, AE-p, and PPPTAE over epochs on the Freeway dataset.}
  \label{Loss}
\end{figure}

\subsubsection{MSE Loss Comparison}

To further intuitively demonstrate the modeling capability of PPPTAE for normal data, we compare the Mean Squared Error (MSE) training loss of PPPTAE with AE\_r and AE\_p. We present the results on the Freeway dataset for brevity, as results on other datasets are similar. The results are shown in Fig. \ref{Loss}, where the loss value corresponding to each epoch is the average of the losses from all mini-batches in the training set. On one hand, for AE\_r, a smaller reconstruction error typically indicates an improvement in the learned feature. As mentioned earlier, the predictive prior has an effect of adjusting the features, so that the features of PPPTAE are better than AE\_r. Consequently, PPPTAE exhibits lower reconstruction error compared to AE\_r. On the other hand, compared to AE\_p, PPPTAE also achieves smaller MSE loss. This is because, compared to AE\_p, PPPTAE can further leverage the observation at timestamp $t$, allowing PPPTAE to integrate more comprehensive information to extract better features and resulting in relatively smaller output errors.

\begin{figure}[]  
  \centering  
  \includegraphics[width=8cm]{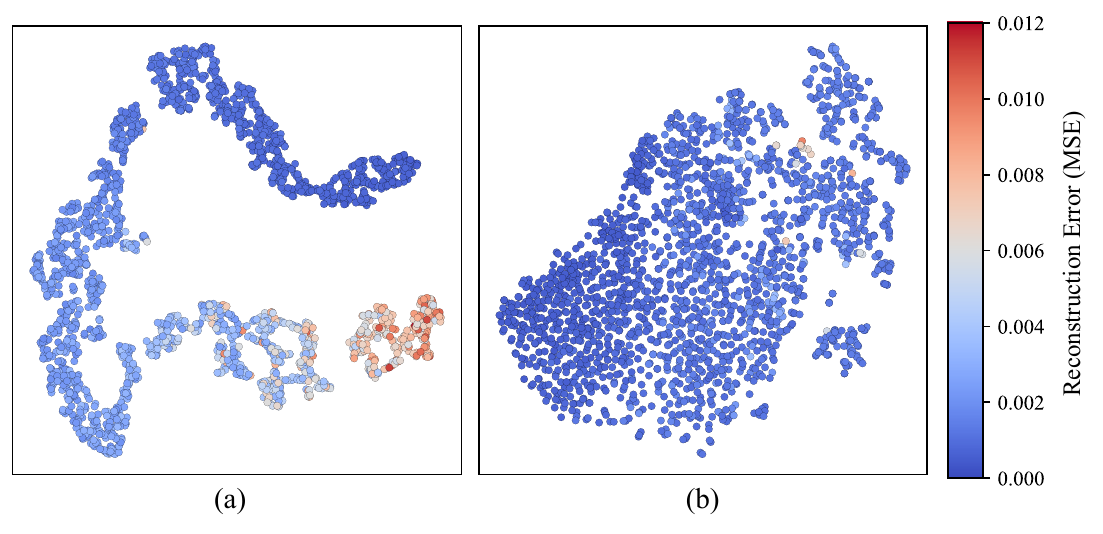} 
  \caption{The feature visualization of (a) AE\_r and (b) PPPTAE on the training set of the Freeway dataset.}  
  \Description{Two t-SNE feature visualizations for AE-r and PPPTAE on the Freeway training set, with colors indicating reconstruction errors.}
  \label{Fea}
\end{figure}

\subsubsection{Feature Visualization}
To further intuitively demonstrate the effect of the predictive prior and the Bayesian fusion approach incorporated in PPPTAE, we use t-SNE \cite{tSNE} to visualize the distribution of the features of PPPTAE on the training set of the Freeway dataset, and compare with that of AE\_r. We also visualized the reconstruction error using colors. The results are shown in Fig. \ref{Fea}. We see from the results that AE\_r exhibits feature clustering, and normal samples are poorly reconstructed in sparse regions, such as that in the lower right side. In contrast, with the predictive prior and the Bayesian fusion approach, the features of PPPTAE have a tendency to be evenly distributed (see Section \ref{Bayes fusion} for details), allowing each region in the feature space to contain sufﬁcient training samples. Consequently, the reconstruction capability corresponding to each region is well established.

\section{Conclusion}

In this work, a Physics-informed Predictive Prior Tensor AE (PPPTAE) framework is proposed for multi-dimensional time series anomaly detection. By incorporating the predictive prior and the Bayesian fusion approach into the reconstruction-based TAE, we bridge the gap between reconstruction-based and prediction-based TAEs, so as to fully leverage the available information and thus the performance is further enhanced. Moreover, by introducing the Bayesian tensor decomposition module, we mitigate the over-generalization problem of AE. In addition, training and testing strategies are proposed for PPPTAE so as to introduce sufficient randomness during training and bypass complex density estimation during testing, respectively. Experimental results demonstrate the eﬀectiveness and superiority of the proposed method.


\bibliographystyle{ACM-Reference-Format}
\bibliography{references}

\appendix

\section*{Appendices}

\section{Probabilistic Modeling} \label{probabilistic Modeling}

For $\mathcal{G}^{(k)}(r_k,r_{k+1},l_k,i_k)$, which is one of the elements in $\mathcal{G}^{(k)}$, we model it using the Gaussian distribution for simplicity and without the loss of generality, i.e.,
\begin{align}
p\left(\mathcal{G}^{(k)}(r_k,r_{k+1},l_k,i_k) \mid \lambda_{r_k}^{(k)},\lambda_{r_{k+1}}^{(k+1)},\lambda_{l_k}^{(k)}\right)= \mathcal{N}\left(\mathcal{G}^{(k)}(r_k,r_{k+1},l_k,i_k) ; 0, \left(\lambda_{r_k}^{(k)} \cdot \lambda_{r_{k+1}}^{(k+1)}\cdot \lambda_{l_k}^{(k)}\right)^{-1}\right), \notag
\end{align}
where the precisions $\lambda_{r_k}^{(k)},\lambda_{r_{k+1}}^{(k+1)}, \lambda_{l_k}^{(k)}$ control the corresponding elements of the TW rank, respectively, $k \in [3]$, and $\lambda_{r_{4}}^{(4)}$ stands for $\lambda_{r_{1}}^{(1)}$. Therefore, we have
\begin{align}
  p\left(\mathcal{G}^{(k)}\mid \boldsymbol{\lambda}^{(k)}_{R}, \boldsymbol{\lambda}^{(k+1)}_{R}, \boldsymbol{\lambda}^{(k)}_{L}\right) =\prod_{r_k=1}^{R_k} \prod_{r_{k+1}=1}^{R_{k+1}} \prod_{l_k =1}^{L_k}  \mathcal{N}\left(\mathcal{G}^{(k)}(r_k,r_{k+1},l_k,:) \mid \mathbf{0},\left(\lambda_{r_k}^{(k)} \cdot \lambda_{r_{k+1}}^{(k+1)}\cdot \lambda_{l_k}^{(k)}\right)^{-1} \mathbf{I}_{I_k} \right), \notag
\end{align}
where $\boldsymbol{\lambda}^{(k)}_{R}$ denotes the vector composed of all $\lambda_{r_k}^{(k)}$, and $\boldsymbol{\lambda}^{(k)}_{L}$ is defined similarly.

Besides, for $\mathcal{C}(l_1,l_2,l_3)$, which is one of the elements in $\mathcal{C}$, we similarly model it using the Gaussian distribution, that is,
\begin{align}
p\left(\mathcal{C}(l_1,l_2,l_3)\mid \lambda_{l_1}^{(1)},\lambda_{l_2}^{(2)},\lambda_{l_3}^{(3)} \right) =  \mathcal{N}\left(\mathcal{C}(l_1,l_2,l_3); 0,\left(\lambda_{l_1}^{(1)}\cdot\lambda_{l_2}^{(2)} \cdot\lambda_{l_3}^{(3)}\right)^{-1} \right), \notag 
\end{align}
where the precisions $\lambda_{l_1}^{(1)},\lambda_{l_2}^{(2)},\lambda_{l_3}^{(3)}$ respectively control the corresponding elements of the TW rank. Hence, we have 
\begin{align}
  p(\mathcal{C} \mid \{\boldsymbol{\lambda}_{L}^{(k)}\}_{k \in [3]}) = \prod_{l_1=1}^{L_1} \prod_{l_2=1}^{L_2} \prod_{l_3=1}^{L_3} \mathcal{N}\left(\mathcal{C}(l_1,l_2,l_3)\mid 0,\left(\lambda_{l_1}^{(1)}\lambda_{l_2}^{(2)} \lambda_{l_3}^{(3)}\right)^{-1} \right). \notag 
\end{align}
Further, to simplify the analysis, we use the conjugate distribution of the Gaussian distribution, namely the Gamma distribution, to model the precision parameters $\boldsymbol{\lambda}_R^{(k)},\boldsymbol{\lambda}_{L}^{(k)}, k \in [3]$, that is,
\begin{align}
  p(\boldsymbol{\lambda}_R^{(k)}) &= \prod_{r_k=1}^{R_k} \operatorname{Ga}\left(\lambda_{r_k}^{(k)} \mid a_{0}^{\lambda_{R}},b_{0}^{\lambda_{R}} \right) \label{lambda_r gamma}\\
  p(\boldsymbol{\lambda}_L^{(k)}) &= \prod_{l_k=1}^{L_k} \operatorname{Ga}\left(\lambda_{l_k}^{(k)} \mid a_{0}^{\lambda_{L}},b_{0}^{\lambda_{L}} \right), \label{lambda_l gamma}
\end{align}
where $\operatorname{Ga}(x \mid a, b)=\frac{b^{a} x^{a-1} e^{-b x}}{\Gamma(a)}$ with $\Gamma(a)$ being the Gamma function, and $a$ and $b$ being the hyper-parameters.

For the sparse component $\mathcal{S}$, considering that it captures the sparse noise and most elements are zero with only a few having large magnitudes, we model it using the independent Gaussian distribution, that is,
\begin{equation}
  p(\mathcal{S} \mid \boldsymbol{\beta}) = \prod_{i_1 =1}^{I_1} \prod_{i_2 =1}^{I_2} \prod_{i_3 =1}^{I_3} \mathcal{N} (\mathcal{S}_{i_1 i_2 i_3} \mid 0,\beta_{i_1 i_2 i_3}^{-1}),
\end{equation}
where $\boldsymbol{\beta}$ is the vector composed of all $\beta_{i_1 i_2 i_3}$, and $\beta_{i_1 i_2 i_3}$ is the precision corresponding to the $(i_1, i_2, i_3)$-th element in $\mathcal{S}$. Further, we use the conjugate distribution of the Gaussian distribution, i.e. the Gamma distribution, to model the precision $\beta_{i_1 i_2 i_3}$, that is,
\begin{equation}
  p(\boldsymbol{\beta}) = \prod_{i_1 =1}^{I_1} \prod_{i_2 =1}^{I_2} \prod_{i_3 = 1}^{I_3} \operatorname{Ga}(\beta_{i_1 i_2 i_3} \mid a_{0}^{\beta},b_{0}^{\beta}). \label{beta gamma}
\end{equation}

For the noise component $\mathcal{E}$, considering that it absorbs the noises except the sparse noise, we model it using the Gaussian distribution and assume that each element in $\mathcal{E}$ shares the same noise precision, i.e.,
\begin{equation}
  p(\mathcal{E} \mid \tau) = \prod_{i_1 =1}^{I_1} \prod_{i_2 =1}^{I_2} \prod_{i_3 = 1}^{I_3}  \mathcal{N}( \mathcal{E}_{i_1 i_2 i_3}\mid 0, \tau^{-1}),
\end{equation}
where $\tau$ is the noise precision. Similar to (\ref{lambda_r gamma}), (\ref{lambda_l gamma}) and (\ref{beta gamma}), we also use the conjugate distribution of the Gaussian distribution, i.e. the Gamma distribution, to model the noise precision $\tau$, that is,
\begin{equation}
  p(\tau) = \operatorname{Ga}(\tau \mid a_{0}^{\tau}, b_{0}^{\tau}).
\end{equation}

Based on Eq.(\ref{decomp}) and the modeling above, given the model parameters, the conditional distribution of the feature $\mathcal{F}$ is
\begin{align}
  &p(\mathcal{F} \mid \{\mathcal{G}^{(k)}\}_{k\in [3]},\mathcal{C},\mathcal{S},\tau) = \prod_{i_1=1}^{I_1} \prod_{i_2=1}^{I_2} \prod_{i_3 =1}^{I_3} \mathcal{N}\left(\mathcal{F}_{i_1 i_2  i_3} \mid \mathrm{TW} [\{\mathcal{G}^{(k)}\}_{k \in [3]} ; \mathcal{C}]_{i_1 i_2 i_3} + \mathcal{S}_{i_1 i_2 i_3},\tau^{-1}\right).  \label{F likelihood}
\end{align}

Therefore, the joint distribution is
{\small
\begin{align}
&p(\mathcal{F},\boldsymbol{\Theta}) =  p(\mathcal{F} \mid \{\mathcal{G}^{(k)}\}_{k \in [3]},\mathcal{C},\mathcal{S},\tau)  \prod_{k=1}^{3} p\left(\mathcal{G}^{(k)}\mid \boldsymbol{\lambda}^{(k)}_{R}, \boldsymbol{\lambda}^{(k+1)}_{R}, \boldsymbol{\lambda}^{(k)}_{L}\right) \notag \\
& \qquad\qquad\times p(\mathcal{C} \mid \{\boldsymbol{\lambda}_{L}^{(k)}\}_{k \in [3]} )\prod_{k=1}^{3} p(\boldsymbol{\lambda}_R^{(k)})  \prod_{k=1}^{3} p(\boldsymbol{\lambda}_L^{(k)})p(\mathcal{S} \mid \boldsymbol{\beta}) p(\boldsymbol{\beta})p(\tau), \notag
\end{align}
}
where $\boldsymbol{\Theta} = \left\{ \{\mathcal{G}^{(k)}\}_{k \in [3]},\mathcal{C},\{\boldsymbol{\lambda}^{(k)}_{R}\}_{k \in [3]},\{\boldsymbol{\lambda}^{(k)}_{L}\}_{k \in [3]}, \mathcal{S},\boldsymbol{\beta},\tau\right\}$ is the set of all variables.

\section{Optimal Variational Posterior Distributions} \label{optimal}

The optimal variational posterior distributions of all the latent variables are given as follows:
\begin{align}
  q(\mathcal{G}^{(k)}(r_k,&r_{k+1},l_k,i_k)) = \mathcal{N}\Biggl(\mathcal{G}^{(k)}(r_k,r_{k+1},l_k,i_k)\mid  \langle\mathcal{G}^{(k)}(r_k,r_{k+1},l_k,i_k)\rangle, \sigma_{g^{k}_{r_k,r_{k+1},l_k,i_k}}^{2}\Biggl)
\end{align}
\begin{align}
q(\mathcal{C}_{l_1 l_2 l_3}) &= \mathcal{N}\left(\mathcal{C}_{l_1 l_2 l_3}\mid\left\langle\mathcal{C}_{l_1 l_2 l_3}\right\rangle, \sigma_{C_{l_1 l_2 l_3}}^{2}\right) \\
q(\lambda_{r_k}^{(k)})&=\mathrm{Ga}\left(\lambda_{r_k}^{(k)}\mid a^{\lambda_{r_k}^{(k)}}, b^{\lambda_{r_k}^{(k)}}\right) \\
q(\lambda_{l_k}^{(k)})&=\mathrm{Ga}\left(\lambda_{l_k}^{(k)}\mid a^{\lambda_{l_k}^{(k)}}, b^{\lambda_{l_k}^{(k)}}\right)\\
q(\tau)&=\mathrm{Ga}\left(\tau \mid a^{\tau}, b^{\tau}\right) \\
q(\mathcal{S}_{i_1 i_2 i_3})&= \mathcal{N}(\mathcal{S}_{i_1 i_2 i_3} \mid \left\langle  \mathcal{S}_{i_1 i_2 i_3}\right\rangle,\sigma^{2}_{S_{i_1 i_2 i_3}}) \\
q(\boldsymbol{\beta}_{i_1 i_2 i_3}) &= \mathrm{Ga} (\boldsymbol{\beta}_{i_1 i_2 i_3} \mid a^{\beta_{i_1 i_2 i_3}},b^{\beta_{i_1 i_2 i_3}} ). 
\end{align}

\section{Close Form Expressions} \label{close form}

For the variable $\boldsymbol{\lambda}_{R}^{(k)}$, it appears only in the first KL divergence term of Eq.($\ref{KL solve}$). According to the variational inference theory, we know that
\begin{align}
q(\lambda_{r_k}^{(k)}) &\propto \exp \left( \langle \ln  p(\mathcal{F},\boldsymbol{\Theta}) \rangle_{\boldsymbol{\Theta} \backslash \lambda_{r_k}^{(k)}} \right).
\end{align}
where $\lambda_{r_k}^{(k)}$ is the $r_k$-th element of $\boldsymbol{\lambda}_{R}^{(k)}$. Therefore, we have
\begin{align}
a^{\lambda_{r_k}^{(k)}}&=a_0^{\lambda_R} +\frac{L_k R_{k+1} I_{k} +L_{k-1}R_{k-1}I_{k-1}}{2}\\
b^{\lambda_{r_k}^{(k)}}&=b_0^{\lambda_R}+\frac{1}{2}\sum_{l_k=1}^{L_{k}}\sum_{r_{k+1}=1}^{R_{k+1}}\sum_{i_k=1}^{I_{k}}\left\langle\lambda_{r_{k+1}}^{(k+1)}\right\rangle \left\langle\lambda_{l_k}^{(k)}\right\rangle \times\left\langle\mathcal{G}^{(k)^2}(r_k,r_{k+1},l_k,i_k)\right\rangle  \notag \\
&\quad + \frac{1}{2}\sum_{l_{k-1}=1}^{L_{k-1}} \sum_{r_{k-1}=1}^{R_{k-1}}\sum_{i_{k-1}=1}^{I_{k-1}}\left\langle\lambda_{r_{k-1}}^{(k-1)}\right\rangle \left\langle\lambda_{l_{k-1}}^{(k-1)}\right\rangle \left\langle\mathcal{G}^{(k-1)^2}(r_{k-1},r_{k},l_{k-1},i_{k-1} )\right\rangle\\
\langle \lambda_{r_k}^{(k)} \rangle &= \frac{a^{\lambda_{r_k}^{(k)}}}{b^{\lambda_{r_k}^{(k)}}}. \label{mean lambda_r}
\end{align}
Similarly, for $q(\lambda_{l_k}^{(k)})$, with $\lambda_{l_k}^{(k)}$ being the $l_k$-th elements of $\boldsymbol{\lambda}_{L}^{(k)}$, we have
\begin{align}
q(\lambda_{l_k}^{(k)})\propto \exp \left( \langle \ln  p(\mathcal{F},\boldsymbol{\Theta}) \rangle_{\boldsymbol{\Theta} \backslash \lambda_{l_k}^{(k)}} \right),
\end{align}
thus we know that
{\small
\begin{align}
&a^{\lambda_{l_k}^{(k)}}=a_0^{\lambda_L} +\frac{R_k R_{k+1} I_{k}+ \prod_{j =1 (\ne k)}^{d} L_j}{2} \notag\\
&b^{\lambda_{l_k}^{(k)}}=b_0^{\lambda_L}+\frac{1}{2}\sum_{r_k =1}^{R_{k}}  \sum_{r_{k+1}=1}^{R_{k+1}} \sum_{i_k =1}^{I_k} \left\langle\lambda_{r_k}^{(k)}\right\rangle \left\langle\lambda_{r_{k+1}}^{(k+1)}\right\rangle \left\langle\mathcal{G}^{(k)^2}(r_k,r_{k+1},l_k,i_k)\right\rangle  \notag \\
&\qquad\qquad+ \frac{1}{2} \sum_{j=1(\ne k)}^{d} \sum_{l_j =1}^{L_j} \left(\prod_{h=1 (\ne k)}^{d}\left\langle\lambda_{l_h}^{(h)}\right\rangle \left\langle\mathcal{C}^{2}(l_1,l_2,\cdots,l_N) \right\rangle \right) \notag
\\
&\langle \lambda_{l_k}^{(k)} \rangle = \frac{a^{\lambda_{l_k}^{(k)}}}{b^{\lambda_{l_k}^{(k)}}}.  \label{mean lambda_l}
\end{align}
}
Similarly, for $q(\tau)$, we can minimize
\begin{align}
&\mathrm{KL}[q(\tau) || p(\tau)] -  \mathbb{E}_{q(\tau)} [\ln p(\mathcal{F} \mid \{\hat{\mathcal{G}}^{(k)}\}_{k \in [3]}, \hat{\mathcal{C}},\hat{\mathcal{S}},\tau )] \notag\\
=&\mathrm{KL}[q(\tau) || p(\mathcal{F} \mid \{\hat{\mathcal{G}}^{(k)}\}_{k \in [3]}, \hat{\mathcal{C}},\hat{\mathcal{S}},\tau )p(\tau)],
\end{align}
so as to obtain the parameters of $q(\tau)$. Therefore, we have 
\begin{align}
q(\tau)\propto \exp \left( \langle \ln  p(\mathcal{F},\boldsymbol{\Theta}) \rangle_{\boldsymbol{\Theta} \backslash \tau} \right),
\end{align}
and thus
\begin{align}
a^{\tau} &= a_0^{\tau} + \frac{I_1 I_2 I_3}{2}\\
b^{\tau} &= b_{0}^{\tau} + \frac{ \left\|   \mathcal{F} - \mathrm{TW} [\{\hat{\mathcal{G}}^{(k)}\}_{k \in [3]} ; \hat{\mathcal{C}}] - \hat{\mathcal{S}} \right\|^{2}_{F} }{2} \\
\langle\tau \rangle &= \frac{a^{\tau}}{b^{\tau}}. \label{mean_tau}
\end{align}

For $q(\beta_{i_1  i_2 i_3})$, with $\beta_{i_1  i_2 i_3}$ being the $(i_1,i_2,i_3)$-th element of $\boldsymbol{\beta}$, we have 
\begin{align}
q(\beta_{i_1  i_2 i_3})&\propto \exp \left( \langle \ln  p(\mathcal{F},\boldsymbol{\Theta}) \rangle_{\boldsymbol{\Theta} \backslash \beta_{i_1  i_2 i_3}} \right),
\end{align}
so that
\begin{align}
a^{\beta_{i_1  i_2 i_3}} &= a_0^{\beta} + \frac{1}{2}\\
b^{\beta_{i_1  i_2 i_3}} &= b_0^{\beta} + \frac{1}{2}\left(\left\langle \mathcal{S}_{i_1  i_2 i_3} \right\rangle^{2} + \sigma^{2}_{S_{i_1  i_2 i_3}} \right)\\
\langle \beta_{i_1  i_2 i_3} \rangle &=\frac{a^{\beta_{i_1  i_2 i_3}}}{b^{\beta_{i_1  i_2 i_3}}} \label{mean_beta}.
\end{align}

\section{Variational Bayes Loss Function} \label{deri loss}

Given $\langle\tau\rangle$, according to the second term in Eq.(\ref{eq.14}), we can obtain $L_{\mathcal{F}}$ in Table \ref{tab:loss VB}.

Besides, given $\langle \beta\rangle$, we can further derive the last KL divergence term in Eq.(\ref{KL solve}), that is,
\begin{align}
& \mathrm{KL}[q(\mathcal{S})q(\boldsymbol{\beta}) || p(\mathcal{S} \mid \boldsymbol{\beta})p(\boldsymbol{\beta})] \notag \\
&=\int q(\mathcal{S})q(\boldsymbol{\beta}) \ln \left( \frac{q(\mathcal{S})q(\boldsymbol{\beta})}{ p(\mathcal{S} \mid \boldsymbol{\beta})p(\boldsymbol{\beta})} \right) d \mathcal{S} d \boldsymbol{\beta} \notag\\
&= \frac{1}{2} \sum_{i_1 i_2 i_3 }  \langle  \beta_{i_1 i_2 i_3} \rangle  \left\langle  \mathcal{S}_{i_1 i_2 i_3}\right\rangle^{2} + \frac{1}{2} \sum_{i_1 i_2 i_3 }\left(   \langle  \beta_{i_1 i_2 i_3} \rangle \sigma^{2}_{S_{i_1 i_2 i_3}} -\ln \sigma^{2}_{S_{i_1 i_2 i_3}}  \right) + \text{const}.
\end{align}

Thus, we can obtain $L_{\langle \mathcal{S}\rangle}$ and $L_{\sigma_{S}^{2}}$ in Table \ref{tab:loss VB}.

Next, given $\langle\boldsymbol{\lambda}_{R}^{(k)}\rangle$ and $\langle \boldsymbol{\lambda}_{L}^{(k)}\rangle$, we can further derive the first KL divergence term in Eq.(\ref{KL solve}). Also, with
{\small
\begin{align}
\mathbb{E}&[\ln q(\mathcal{G}^{(k)}(r_k,r_{k+1},l_k,i_k))] =- \frac{1}{2} \ln  \sigma_{g^{k}_{r_k,r_{k+1},l_k,i_k}}^{2} + \text{const} \notag \\
\mathbb{E}&[\ln q(\mathcal{C}_{l_1\cdots l_d})] = -\frac{1}{2} \ln   \sigma_{C_{l_1\cdots l_d}}^{2} + \text{const} \notag\\
\mathbb{E}&[\ln p(\mathcal{G}^{(k)}(r_k,r_{k+1},l_k,i_k)\mid \lambda_R,\lambda_L)] = - \frac{1}{2}\langle \lambda_{r_k}^{(k)} \rangle\langle  \lambda_{r_{k+1}}^{(k+1)} \rangle \langle \lambda_{l_k}^{(k)} \rangle \Bigl(  \left.\left\langle\mathcal{G}^{(k)}(r_k,r_{k+1},l_k,i_k)\right\rangle^{2}+ \sigma_{g^{k}_{r_k,r_{k+1},l_k,i_k}}^{2}  \right) \notag \\
\mathbb{E}&[\ln p(\mathcal{C}_{l_1,l_2, l_3}\mid \lambda_L)] = -\frac{1}{2} \langle \lambda_{l_1}^{(1)} \rangle \langle \lambda_{l_2}^{(2)} \rangle \langle \lambda_{l_3}^{(3)}\rangle \left( \left\langle\mathcal{C}_{l_1 l_2 l_3}\right\rangle^{2} + \sigma_{C_{l_1 l_2 l_3}}^{2}\right), \notag
\end{align}
}

we can obtain the $L_{\langle \mathcal{G}^{(k)}\rangle}$, $L_{\sigma_{g^{k}}^{2}}$, $L_{\langle  \mathcal{C}  \rangle}$, and $L_{\sigma_{C}^{2}}$ in Table \ref{tab:loss VB}.

So far, we have obtained all the loss functions listed in Table \ref{tab:loss VB}.

\end{document}